\documentclass[letterpaper]{article} % DO NOT CHANGE THIS
\usepackage{aaai2026}  % DO NOT CHANGE THIS
\usepackage{times}  % DO NOT CHANGE THIS
\usepackage{helvet}  % DO NOT CHANGE THIS
\usepackage{courier}  % DO NOT CHANGE THIS
\usepackage[hyphens]{url}  % DO NOT CHANGE THIS
\usepackage{graphicx} % DO NOT CHANGE THIS
\usepackage{natbib}  % DO NOT CHANGE THIS AND DO NOT ADD ANY OPTIONS TO IT
\usepackage{caption} % DO NOT CHANGE THIS AND DO NOT ADD ANY OPTIONS TO IT
\usepackage{algorithm}
\usepackage{algorithmic}

\usepackage{booktabs}
\usepackage{multirow}
\usepackage{makecell}
\usepackage{subcaption}
\usepackage{amssymb}
\usepackage{amsmath}

\usepackage{fancyhdr}     % 引入fancyhdr宏包
\usepackage{newfloat}
\usepackage{listings}
\DeclareCaptionStyle{ruled}{labelfont=normalfont,labelsep=colon,strut=off} % DO NOT CHANGE THIS
\floatstyle{ruled}
\newfloat{listing}{tb}{lst}{}
\floatname{listing}{Listing}
\title{I2E: Real-Time Image-to-Event Conversion for High-Performance Spiking Neural Networks}
\author{
    Ruichen Ma\textsuperscript{\rm 1},
    Liwei Meng\textsuperscript{\rm 1},
    Guanchao Qiao\textsuperscript{\rm 1},
    Ning Ning\textsuperscript{\rm 1},
    Yang Liu\textsuperscript{\rm 1},
    Shaogang Hu\textsuperscript{\rm 1,\rm 2,}\thanks{Corresponding author.}
}
\affiliations{
    \textsuperscript{\rm 1}School of Integrated Circuit Science and Engineering, University of Electronic Science and Technology of China\\
    \textsuperscript{\rm 2}Shenzhen Institute for Advanced Study, University of Electronic Science and Technology of China \\
    \{ruichen.ma, lw\_meng\}@std.uestc.edu.cn, \{gcqiao, ning\_ning, yliu1975, sghu\}@uestc.edu.cn
}

\usepackage{bibentry}
\begin{document}

\maketitle
\thispagestyle{plain}
\begin{abstract}

Spiking neural networks (SNNs) promise highly energy-efficient computing, but their adoption is hindered by a critical scarcity of event-stream data.
This work introduces I2E, an algorithmic framework that resolves this bottleneck by converting static images into high-fidelity event streams.
By simulating microsaccadic eye movements with a highly parallelized convolution, I2E achieves a conversion speed over 300x faster than prior methods, uniquely enabling on-the-fly data augmentation for SNN training.
The framework's effectiveness is demonstrated on large-scale benchmarks.
An SNN trained on the generated I2E-ImageNet dataset achieves a state-of-the-art accuracy of 60.50\%.
Critically, this work establishes a powerful sim-to-real paradigm where pre-training on synthetic I2E data and fine-tuning on the real-world CIFAR10-DVS dataset yields an unprecedented accuracy of 92.5\%.
This result validates that synthetic event data can serve as a high-fidelity proxy for real sensor data, bridging a long-standing gap in neuromorphic engineering.
By providing a scalable solution to the data problem, I2E offers a foundational toolkit for developing high-performance neuromorphic systems.
The open-source algorithm and all generated datasets are provided to accelerate research in the field.

\end{abstract}

\begin{links}
    \link{Code \& Datasets}{https://github.com/Ruichen0424/I2E}
\end{links}

%==================================================== Intro ============================================================================
\section{Introduction}
\label{sec:intro}

\begin{table}[t]
	\centering
	\begin{tabular}{p{6.7em} p{5.7em}<{\centering} p{4.3em}<{\centering} p{2.0em}<{\centering}}
		\toprule[1.5pt]
		{Dataset} & {Architecture} & {Method} & {Acc.\%} \\
		\midrule[1pt]
    % \multirow{2}{*}{ImageNet} & ResNet18+LIF & baseline & 39.89\\
		%  & ResNet18+LIF & pre-train & 43.74 \\
    %  \midrule[1pt]
		\multirow{5}{*}{\makecell{ES-ImageNet \\ \cite{lin2021imagenet}}} & ResNet18+LIF & baseline & 39.89\\
		 & ResNet18+LIF & pre-train & 43.74 \\
		 & ResNet18+LIAF & pre-train & 52.25 \\
		 & ResNet34+LIF & baseline & 43.42 \\
		 & ResNet34+LIAF & pre-train & 51.83 \\
		\midrule[1pt]
		\multirow{4}{*}{\makecell{N-ImageNet \\ \cite{kim2021n}}} & ResNet34 & EH & 47.73\\
		 & ResNet34 & STS & 47.90\\
		 & ResNet34 & DiST & 48.43\\
		 & ResNet34 & EST & 48.93\\
		\midrule[1pt]
		\multirow{4}{*}{\makecell{\textbf{I2E-ImageNet} \\ \textbf{(This work)}}} & ResNet18+LIF & baseline-I & 48.30\\
		 & ResNet18+LIF & baseline-II & 57.97 \\
		 & ResNet18+LIF & pre-train & \textbf{59.28} \\
		 & ResNet34+LIF & baseline-II & \textbf{60.50} \\
		\bottomrule[1.5pt]
	\end{tabular}
	\caption{State-of-the-art comparison on event-based ImageNet classification.
  The proposed I2E-ImageNet enables an MS-ResNet34 architecture to achieve a new state-of-the-art accuracy of 60.50\%, substantially outperforming all prior results and demonstrating the superior quality of the synthetic data for training deep high-performance SNNs.}
	\label{table_imagenet}
\end{table}
\begin{figure*}[t]
  \centering
  \includegraphics[width=\linewidth]{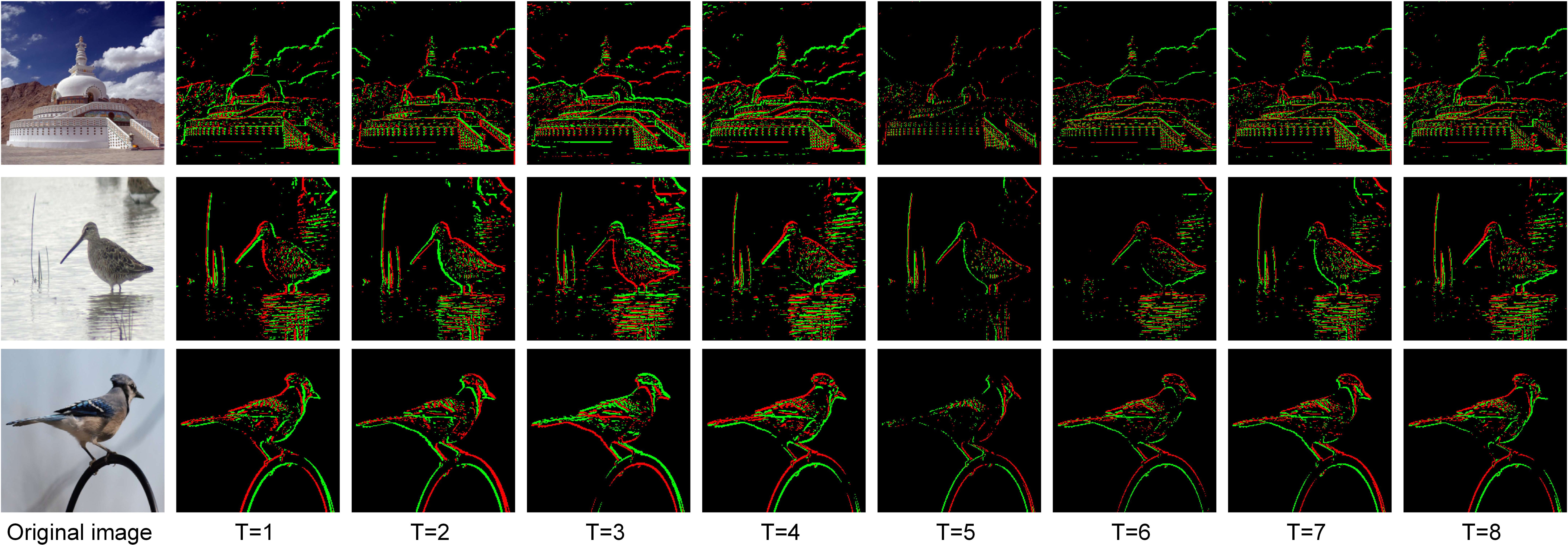}
  \caption{The I2E image-to-event conversion process.
  A single static RGB image is transformed into an eight-timestep event stream by simulating microsaccadic eye movements.
  The process effectively captures fine-grained details and salient object contours, producing a sparse data format well-suited for efficient, event-driven processing by SNNs.}
  \label{fig_ex}
\end{figure*}

Spiking neural networks (SNNs) represent a promising computational paradigm inspired by the brain's sparse, event-driven processing principles \cite{xu2018csnn,zenke2021visualizing}.
This bio-inspired design offers a path toward exceptional energy efficiency \cite{roy2019towards, pei2019towards, zhang2020system, subbulakshmi2021biomimetic}.
When deployed on specialized neuromorphic hardware like Loihi \cite{davies2018loihi} or TrueNorth \cite{merolla2014million}, SNNs can achieve orders-of-magnitude gains in power efficiency over conventional artificial neural networks (ANNs), making them ideal candidates for deployment on power-constrained edge devices.

The natural input for an SNN is a stream of asynchronous events, data typically captured by specialized hardware such as dynamic vision sensors (DVS).
Unlike conventional cameras that record dense frames at fixed intervals, DVS cameras report pixel-level brightness changes as they occur \cite{wu2024recent}.
However, this reliance on specialized hardware has created a fundamental data bottleneck that severely impedes the development and adoption of SNNs.
The acquisition of large-scale event datasets is a resource-intensive and time-consuming process, resulting in benchmarks that are limited in scale.
Furthermore, the quality of existing datasets can be compromised by capture artifacts, such as monitor flicker \cite{serrano2015poker}.
The combination of data scarcity and inconsistent quality has led to a persistent performance gap.
As shown in Table~\ref{table_imagenet}, the accuracy of state-of-the-art networks on event-based ImageNet datasets lags significantly behind their ANN counterparts, with an accuracy of over 70\%, casting doubt on their readiness for complex, real-world applications.

To circumvent this data limitation, a common practice involves repeatedly presenting the same image to an SNN at each timestep \cite{deng2022temporal, zhou2022spikformer, meng2023towards, jiang2024tab, yao2024spike}.
This approach, however, represents a significant compromise, forcing dense, redundant computations that undermine the event-driven paradigm, negating the very energy and latency advantages that make SNNs a compelling alternative to ANNs.
This has created an intractable dilemma for the field: either rely on scarce, low-performing real event data or abandon the core principles of neuromorphic computing.

To resolve this challenge, this paper introduces I2E, an ultra-efficient algorithmic framework that converts the vast repository of static images into high-quality event streams in real-time.
I2E bridges the gap between large-scale image datasets and the data requirements of high-performance SNNs, enabling them to be trained at scale without compromising their fundamental operating principles.
The primary contributions of this work are threefold:
\begin{itemize}
\item An algorithmic framework for real-time image-to-event conversion is presented.
Its processing speed is over 300x faster than prior methods and up to 30,000x faster than physical acquisition, which for the first time enables the use of on-the-fly data augmentation for SNN training.
\item Large-scale, high-quality event-stream datasets, I2E-ImageNet and I2E-CIFAR, are generated.
An SNN trained on I2E-ImageNet achieves 60.50\% accuracy, establishing a new state-of-the-art for event-based ImageNet and significantly closing the performance gap.
\item A highly effective sim-to-real training paradigm is established.
By pre-training on synthetic I2E-CIFAR10 data and fine-tuning on the real-world CIFAR10-DVS dataset, an unprecedented accuracy of 92.5\% is achieved, demonstrating that I2E-generated data serves as a high-fidelity proxy for real sensor data.
\end{itemize}

The I2E conversion process, illustrated in Figure~\ref{fig_ex}, effectively preserves crucial visual information within a sparse data format.
By open-sourcing the algorithm and the accompanying datasets, this work provides the research community with an essential toolkit to overcome the long-standing data bottleneck, thereby accelerating the development of practical, high-performance neuromorphic systems.

%==================================================== Related ============================================================================
\section{Related Work}
\label{sec:related}

\begin{figure*}[t]
  \centering
  \includegraphics[width=0.85\linewidth]{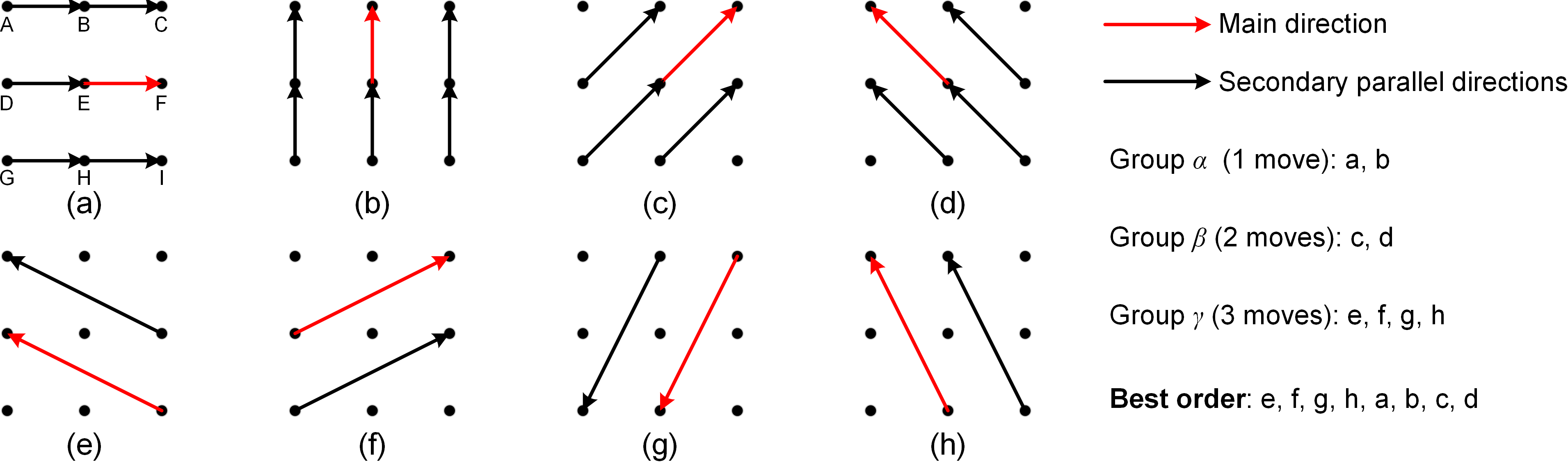}
  \caption{The I2E algorithm simulates microsaccadic eye movements using a source image (point E) and its eight one-pixel-shifted versions (the other points), represented by a $3\times3$ grid.
  The intensity change $\Delta V$ is calculated by differencing pairs of these images.
  As shown by the arrows, these differences are classified into eight directional groups, each of which generates the data for one of the eight timesteps in the final event stream.}
  \label{fig_direction}
\end{figure*}

The acquisition of large-scale, high-quality event-stream data remains a primary obstacle to advancing research on SNNs.
Existing approaches to data generation can be broadly categorized into hardware-based capture and algorithmic conversion, each presenting significant drawbacks that have constrained progress in the field.

The most direct approach to data generation involves using DVS to capture events.
One strategy is to record real-world scenes, which produces data with high temporal fidelity.
However, this process is resource-intensive and slow, resulting in datasets that are often limited in scale and scope, such as DVS-Gesture \cite{amir2017low} and DailyAction \cite{liu2021event}, or others focused on specific object or action categories \cite{miao2019neuromorphic, bi2019graph, bi2020graph, wang2021event, vasudevan2022sl, dong2023bullying10k, wang2024dailydvs, sironi2018hats, bolten2021dvs}.
The limited sample sizes of these datasets are often insufficient for training the deep SNN architectures required for complex recognition tasks.

To address the issue of scale, an alternative hardware-based strategy involves recording a monitor that displays static images, leading to widely used benchmarks like N-MNIST \cite{orchard2015converting}, CIFAR10-DVS \cite{li2017cifar10}, and N-ImageNet \cite{kim2021n}.
Although this method increases the number of available samples, it introduces distinct challenges.
The resulting datasets may suffer from significant data degradation due to capture artifacts, such as LCD screen flicker.
Moreover, the acquisition process remains exceedingly slow.
Generating the N-ImageNet dataset, for instance, required several days of continuous recording, rendering the process impractical for expansion or modification.
A further complication of both hardware-based approaches is the ultra-high temporal resolution of the raw data, which often must be integrated into a smaller number of timesteps for practical use in SNNs \cite{fang2021incorporating}.
This integration process can compromise the sparse, binary nature of the event stream.

To bypass the limitations of physical hardware, several algorithmic methods have been developed to convert conventional visual data into event streams.
These include techniques for converting video \cite{bi2017pix2nvs, gehrig2020video, hu2021v2e} and, more recently, static images.
A notable example is the ODG algorithm used to create the ES-ImageNet dataset \cite{lin2021imagenet}.
While algorithmic conversion drastically reduces cost, existing methods are critically hampered by a computational bottleneck.
The generation of a large-scale dataset like ImageNet can take over ten hours on modern hardware.
This severe latency makes these algorithms unsuitable for real-time applications and, crucially, precludes the use of on-the-fly data augmentation, a standard and vital technique for training state-of-the-art neural networks.

The I2E framework is designed to resolve these trade-offs.
It sidesteps the speed, cost, and quality limitations of hardware acquisition while overcoming the computational bottleneck of prior algorithmic methods, providing a scalable, cost-effective, and practical foundation for training high-performance SNNs within modern deep learning workflows.

%============================================================= Method ==========================================================
\section{Method}

This section details the I2E algorithm, a framework for generating high-fidelity event streams from static images in real time.
The exposition first presents the core mechanics of the conversion pipeline, followed by a theoretical analysis that quantifies the algorithm's advantages in terms of speed, energy cost, and information compression.

\begin{algorithm}[tb]
\caption{The I2E Conversion Algorithm}
\label{alg_i2e}
\textbf{Input}: A batch of RGB images $I \in \mathbb{R}^{B \times 3 \times H \times W}$. \\
\textbf{Output}: A batch of binary spikes $S \in \mathbb{B}^{T \times B \times 2 \times H \times W}$. \\
\begin{algorithmic}[1] %[1] enables line numbers
\STATE // \textit{Define 8 kernels for 8 motion directions (timesteps)}
\STATE $v \leftarrow [[9,4], [4,3], [3,8], [8,1], [5,6], [5,2], [5,3], [5,1]]$
\STATE $K \leftarrow \text{zeros}(8, 1, 3, 3)$
\FOR{$t \in [0, 7]$}
    \STATE $y_0, x_0 \leftarrow (v[t][0]-1)//3, (v[t][0]-1)\%3$
    \STATE $y_1, x_1 \leftarrow (v[t][1]-1)//3, (v[t][1]-1)\%3$
    \STATE $K[t, 0, y_0, x_0] \leftarrow -1$; $K[t, 0, y_1, x_1] \leftarrow 1$
\ENDFOR
\STATE // \textit{Convert RGB to intensity and compute changes}
\STATE $V \leftarrow \max(I)$
\STATE $\Delta V \leftarrow \text{conv2d}(\text{pad}(V, 1), K)$
\STATE // \textit{Apply dynamic threshold for event generation}
\STATE $V_{range} \leftarrow \max(V) - \min(V)$
\STATE $S_{th} \leftarrow S_{th_0} \cdot V_{range}$
\STATE $S_{ON} \leftarrow (\Delta V > S_{th}).\text{float}()$ \hfill \textit{\# ON events}
\STATE $S_{OFF} \leftarrow (-\Delta V > S_{th}).\text{float}()$ \hfill \textit{\# OFF events}
\STATE $S \leftarrow \text{stack}([S_{ON}, S_{OFF}], \text{dim}=2)$
\STATE \textbf{return} $S.\textit{permute}(1,0,2,3,4)$
\end{algorithmic}
\end{algorithm}

\subsection{The I2E Conversion Pipeline}
The I2E algorithm (pseudocode see Algorithm~\ref{alg_i2e}) transforms a static image into a temporally dynamic event stream through three key stages.
The entire pipeline is designed as a sequence of highly parallelizable tensor operations, making it ideally suited for GPU acceleration.

\subsubsection{Stage 1: Intensity Map Generation}
A DVS camera responds to changes in logarithmic brightness.
To efficiently emulate this, a standard RGB image $I_{RGB} \in \mathbb{R}^{3 \times H \times W}$ is first converted into a single-channel intensity map $V \in \mathbb{R}^{1 \times H \times W}$.
For this purpose, the Value (V) channel from the HSV color space is used, as it represents the maximum intensity across the R, G, and B channels and can be extracted with negligible computational cost according to Equation~\ref{eq:value}.
\begin{equation}
  V(x, y) = \max(I_R(x,y), I_G(x,y), I_B(x,y))
  \label{eq:value}
\end{equation}
This choice prioritizes speed while producing an intensity representation analogous to the information captured by a sensor's photoreceptors.
The performance impact of color-to-grayscale conversion is quantified in the ablation study.

\begin{figure}[t]
  \centering
  \begin{subfigure}{0.49\linewidth}
    \centering
    \includegraphics[width=0.75\linewidth]{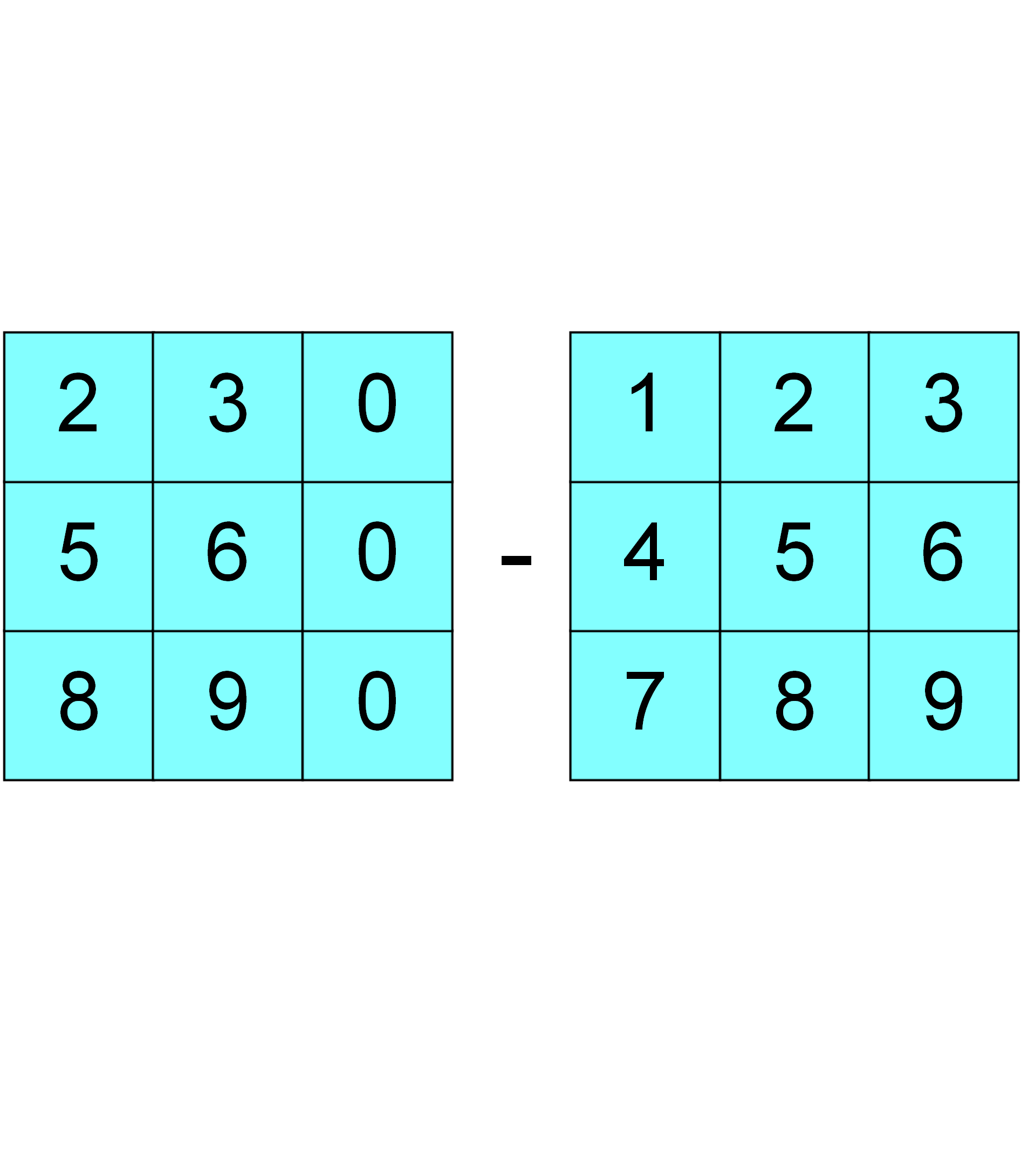}
    \caption{Image differencing.}
    \label{fig2a}
  \end{subfigure}
  \hfill
  \begin{subfigure}{0.49\linewidth}
    \centering
    \includegraphics[width=0.65\linewidth]{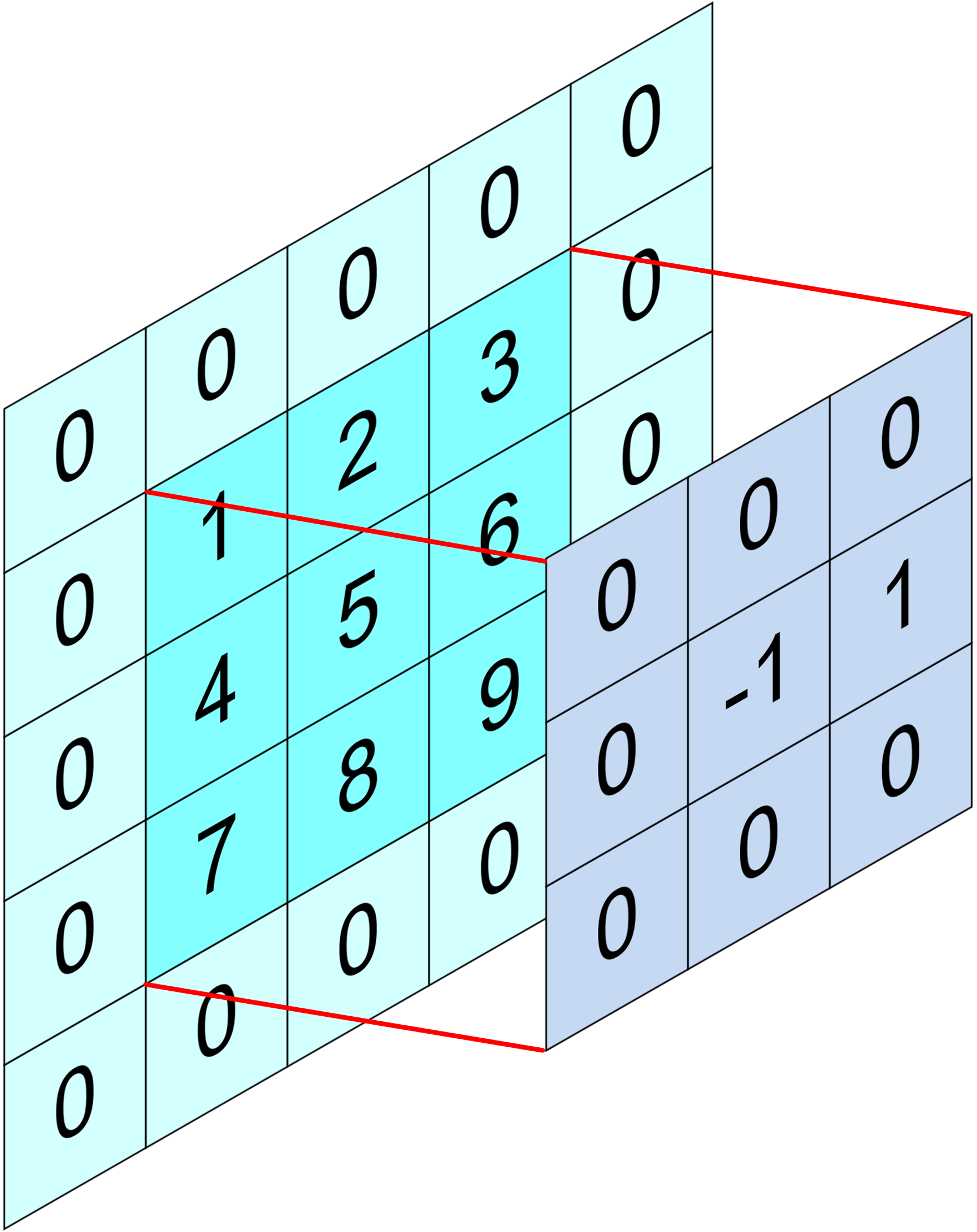}
    \caption{Equivalent convolution.}
    \label{fig2b}
  \end{subfigure}
  \caption{The subtraction of two shifted images is computationally equivalent to a 2D convolution with a sparse kernel.}
  \label{fig2}
\end{figure}

\subsubsection{Stage 2: Event Generation via Spatio-Temporal Convolution}
A central innovation of I2E is its method for generating temporal dynamics from the static intensity map.
The algorithm simulates the effect of microsaccadic eye movements, small, involuntary saccades, by calculating the difference between slightly shifted versions of the intensity map.
A naive implementation would require multiple memory-intensive image translation and subtraction operations.
Instead, I2E implements this process as a single, highly efficient 2D convolution, as illustrated in Figure~\ref{fig2}.

Translating the image by a single pixel in any of the eight directions produces a set of nine images (including the original).
The various possible one-pixel motion vectors are classified into eight directional groups, as shown in Figure~\ref{fig_direction}.
The difference between pairs of these images simulates the intensity change $\Delta V$ that a DVS would capture over a short time interval.
For each timestep, a unique $3\times3$ kernel $K_t$ is constructed.
Each kernel is extremely sparse, containing only a single -1 and a single +1 at positions corresponding to the start and end points of the simulated motion vector.
The full set of eight intensity-change maps $\Delta V$ is then generated in a single, parallel operation, as Equation \ref{eq:convolution}.
\begin{equation}
    \Delta V_t = V * K_t
    \label{eq:convolution}
\end{equation}
This formulation is critical for the algorithm's real-time performance on GPUs.
Ablation study reveals an optimal processing sequence for these groups.

To enhance robustness, a stochastic augmentation strategy is employed during training.
For each of the eight directions, a set of equivalent one-pixel shift vectors is defined.
During training, one vector is randomly selected from its set to construct the kernel $K_t$, introducing diversity with no additional computational cost.
For inference, a fixed, canonical vector from each set is used to ensure deterministic output.

\subsubsection{Stage 3: Adaptive Event Firing}
\begin{figure}[t]
  \centering
  \includegraphics[width=0.85\linewidth]{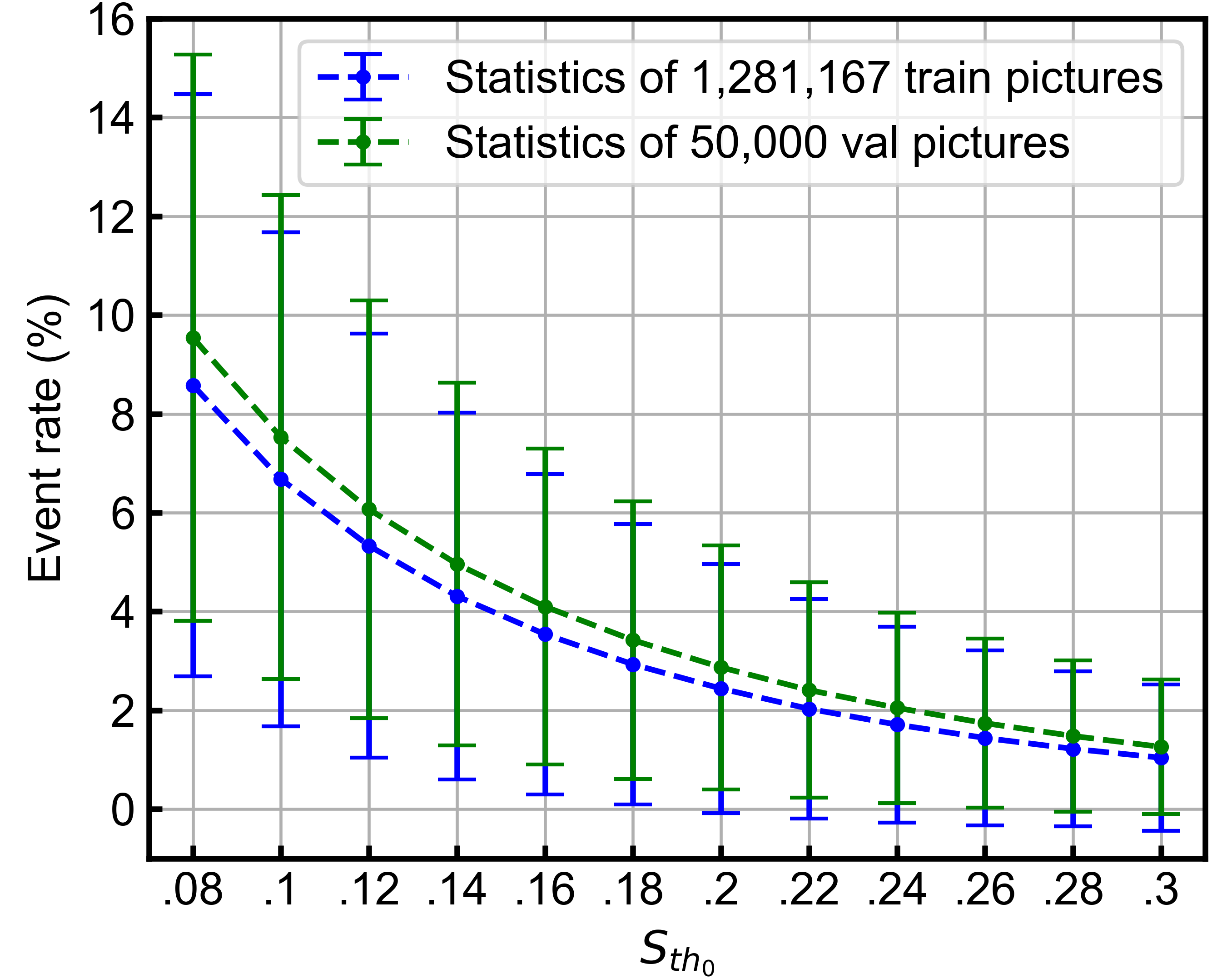}
  \caption{Event rate statistics on ImageNet.
          $S_{th_0} = 0.12$ is selected to achieve a mean event rate of approximately 5\%.}
  \label{fig_sth}
\end{figure}

The final stage converts the continuous intensity-change maps $\Delta V$ into binary spike events.
A pixel at position $(x,y)$ fires an \textit{ON} event if $\Delta V(t,x,y)$ exceeds a positive threshold and an \textit{OFF} event if it falls below a negative threshold, as shown in Equation~\ref{eq:v2s}, where $S \in \mathbb{B}^{8 \times 2 \times H \times W}$.
\begin{equation}
	\left\{\begin{array}{ll}
		S_{(t,0,x,y)} = 1,  &\Delta V_{(t,x,y)}>S_{th}\\
		S_{(t,1,x,y)} = 1, &\Delta V_{(t,x,y)}<-S_{th}\\
		S_{(t,p,x,y)} = 0, & {otherwise.}
		\end{array}
	\right.
	\label{eq:v2s}
\end{equation}
A fixed, global threshold is suboptimal, as it produces inconsistent event rates for images with varying brightness.
Therefore, I2E employs a dynamic threshold $S_{th}$ that adapts to each image's content as shown in Equation \ref{eq:dyn_thresh}.
\begin{equation}
    S_{th} = S_{th_0} \cdot (\max(V) - \min(V))
    \label{eq:dyn_thresh}
\end{equation}
where $S_{th_0}$ is a single global sensitivity hyperparameter.
This adaptive mechanism ensures a more consistent event sparsity across the dataset, which is critical for robust SNN training.
The parameter $S_{th_0}$ directly controls the overall event rate.
Figure~\ref{fig_sth} presents the resulting event rate statistics across the ImageNet dataset for a range of $S_{th_0}$ values.
To balance information preservation and computational efficiency, $S_{th_0}$ is selected to achieve a specific target event rate \cite{lin2021imagenet}.
For ImageNet, $S_{th_0} = 0.12$ is used to achieve a target event rate of approximately 5\%, while for CIFAR datasets, $S_{th_0}$ is set to $0.07$.

\begin{table*}[t]
	\centering
	\begin{tabular}{p{7.0em} p{7.5em}<{\centering} p{3.8em}<{\centering} p{7.8em}<{\centering} p{4.6em}<{\centering}}
		\toprule[1.5pt]
		{Dataset} & \makecell{Generation speed \\ (ms/sample)} & {Resolution} & {\# of samples} & {\# of classes} \\
		\midrule[1pt]
		N-CARS & 100 & $80\times40$ & 24,029 & 2 \\
		Poker-DVS & - & $32\times32$ & 131 & 4 \\
		Bullying10K & 2,000 - 20,000 & $346\times260$ & 10,000 & 10 \\
		DVS-Gesture & 6,000 & $128\times128$ & 1,342 & 11 \\
		DVS-OUTLAB & 500 & $768\times512$ & 47,000 & 11 \\
		DailyDVS-200 & 1,000 - 13,000 & $320\times240$ & 22,046 & 200 \\
		\midrule[1pt]
		N-MNIST & 300 & $28\times28$ & 60,000 + 10,000 & 10 \\
		MNIST-DVS & 2,000 - 4,000 & $128\times128$ & 30,000 & 10 \\
		CIFAR10-DVS & 1,200 & $512\times512$ & 10,000 & 10 \\
		DVS-UCF-50 & 6800 & $240\times180$ & 6,676 & 50 \\
		DVS-Caltech101 & 300 & $302\times245$ & 8,709 & 100 \\
		N-ImageNet & - & $224\times224$ & 1,781,167 & 1,000 \\
		N-Omniglot & 4,000 & $346\times260$ & 32,460 & 1,623 \\
		\midrule[1pt]
		ES-ImageNet & 29.47 & $224\times224$ & 1,257,035 + 49,881 & 1,000 \\
		\midrule[1pt]
		{I2E-CIFAR10} & \textbf{0.03} (GPU) & {$128\times128$} & {50,000 + 10,000} & {10} \\
		{I2E-CIFAR100} & \textbf{0.03} (GPU) & {$128\times128$} & {50,000 + 10,000} & {100} \\
		{I2E-ImageNet} & \textbf{0.1} (GPU) & {$224\times224$} & {1,281,167 + 50,000} & {1,000} \\
		\bottomrule[1.5pt]
	\end{tabular}
	\caption{Comparison of event-based dataset generation speeds.
  The proposed I2E algorithm demonstrates orders-of-magnitude faster per-sample generation speed compared to both hardware-based acquisition methods and previous algorithmic approaches.
  Hardware speeds reflect physical capture time, while algorithmic speeds reflect computation time.}
	\label{table_dvscom}
\end{table*}

\subsection{Efficiency and Information Analysis}
The I2E's design yields significant, quantifiable advantages in computational efficiency and information compression.

\subsubsection{Computational, Energy, and Storage Efficiency}
The convolution-based design of I2E enables unprecedented conversion speed.
As shown in Table~\ref{table_dvscom}, I2E processes an image on a modern GPU in approximately 0.1 ms.
This is orders of magnitude faster than both hardware-based acquisition (e.g., \textgreater 30,000x faster than typical DVS camera capture) and prior algorithmic methods like ODG (\textgreater 300x faster), which require a day to process the full ImageNet dataset.
This real-time capability is a critical advance, enabling the seamless integration of I2E into modern training pipelines that rely on on-the-fly augmentation.

This efficiency translates to substantial energy savings.
The energy for a standard ANN convolution is dominated by multiply-accumulate (MAC) operations, $E_{ANN} = N_{ops} \cdot E_{MAC}$.
For a ResNet-style first layer, assuming a 45nm process where a 32-bit floating-point MAC costs $E_{MAC}=4.6$ pJ \cite{horowitz20141}, the energy consumption is approximately $543$ $\mu\text{J}$.
In contrast, the I2E encoding itself is highly efficient ($E_{I2E} \approx 0.36 \mu\text{J}$).
The SNN layer performs only sparse additions, with energy proportional to the event rate $fr$ and timesteps $T$, $E_{SNN} = N_{ops} \cdot E_{AC} \cdot T \cdot fr$ and $E_{AC}=0.9$ pJ.
For the I2E-SNN, the cost is approximately $28.68$ $\mu\text{J}$, representing a ${18.9\times}$ reduction in first-layer energy consumption compared to the standard ANN approach.

Furthermore, the resulting event-stream data is highly compressible.
The I2E-ImageNet dataset, stored as boolean arrays, occupies 47 GB, a $67.8\%$ reduction from the 146 GB of the original JPEG-compressed ImageNet.

\subsubsection{Information-Theoretic Analysis}
To analyze the trade-off between data compression and information preservation, the Shannon entropy of various data representations was computed across the ImageNet dataset.
The entropy $H$ of a data source $X$ with discrete symbols $x_i$ and probabilities $p(x_i)$ is given by Equation~\ref{eq:entropy}, and measures the average information content per symbol (or pixel).
\begin{equation}
    H(X) = - \sum_{i} p(x_i) \log p(x_i)
    \label{eq:entropy}
\end{equation}
The original grayscale images and the single-channel Value images contain nearly identical information content, with average entropies of $7.12 \pm 0.73$ and $7.14 \pm 0.76$, respectively.
In stark contrast, the final I2E event stream, with a typical event rate of 5\%, has a significantly lower entropy of just $1.53 \pm 0.60$.
This indicates that the I2E conversion achieves substantial information compression, retaining less than 22\% of the original data's entropy.
Despite this massive reduction, the empirical performance degradation observed in experiments is comparatively minor.
This outcome strongly suggests that the majority of entropy in static images corresponds to redundant information (such as uniform textures and backgrounds) and validates that the I2E is highly effective at isolating and preserving the sparse, salient features essential for complex recognition tasks.

%============================================================= Experiment ==========================================================
\section{Experiments}
\label{sec:experiments}
This section empirically validates the effectiveness of the I2E framework.
The experiments are designed to assess three key aspects:
1) the performance of SNNs trained on I2E-generated datasets;
2) the transferability of models pre-trained on I2E data to tasks involving real-world neuromorphic sensor data, establishing a new sim-to-real paradigm;
and 3) the impact of the algorithm's specific design choices through a series of ablation studies.

\subsection{Experimental Setup}
\paragraph{Datasets and architectures}
Performance is evaluated on three standard image recognition benchmarks: CIFAR-10/100 \cite{krizhevsky2009learning} and ImageNet \cite{russakovsky2015imagenet}.
The corresponding event-based datasets (I2E-CIFAR and I2E-ImageNet) are generated using the I2E algorithm.
For these datasets, ImageNet images are resized to $224\times224$, while CIFAR images are resized to $128\times128$ to match the resolution of CIFAR10-DVS.
For the sim-to-real evaluation, the CIFAR10-DVS dataset \cite{li2017cifar10} is used.
All experiments employ MS-ResNet architectures \cite{hu2024advancing} with LIF neurons \cite{fang2021deep}.

\paragraph{Implementation details}
Models are trained using the SpikingJelly framework \cite{fang2023spikingjelly} with mixed-precision on two NVIDIA RTX 4090 GPUs.
The training employs a cross-entropy loss with label smoothing ($\epsilon=0.1$) and the SGD optimizer.
Models for CIFAR and ImageNet are trained for 256 and 128 epochs, respectively.
The initial learning rate is set to 0.1, with weight decay of 2e-4 for CIFAR and 1e-5 for ImageNet.

\paragraph{Data formats}
The generated I2E datasets are provided in two formats: dense boolean tensors, convenient for direct loading into deep learning frameworks, and sparse coordinate lists, which are highly compressed and suitable for applications that process events individually.

\begin{table}[t]
    \centering
    \begin{tabular}{p{6.0em} p{6.6em}<{\centering} p{5.2em}<{\centering} p{1.7em}<{\centering}}
        \toprule[1.5pt]
        {Dataset} & {Architecture} & {Method} & {Acc.\%} \\
        \midrule[1pt]
        \multirow{6}{*}{CIFAR10-DVS} & ResNet18 & baseline & 65.6 \\
         & ResNet20 & baseline & 75.56 \\
         & ResNet34 & transfer  & 73.72 \\
         & SpikingResformer & transfer & 84.8 \\
        \cline{2-4}
         & ResNet18 & transfer-I & \textbf{83.1} \\
         & ResNet18 & transfer-II & \textbf{92.5} \\
        \midrule[1pt]
        \multirow{3}{*}{I2E-CIFAR10} & ResNet18 & baseline-I & 85.07 \\
         & ResNet18 & baseline-II & 89.23 \\
         & ResNet18 & transfer-I & \textbf{90.86} \\
        \midrule[1pt]
        \multirow{3}{*}{I2E-CIFAR100} & ResNet18 & baseline-I & 51.32 \\
         & ResNet18 & baseline-II & 60.68 \\
         & ResNet18 & transfer-I & \textbf{64.53} \\
        \bottomrule[1.5pt]
    \end{tabular}
    \caption{Performance on CIFAR datasets.
  Transfer-I denotes fine-tuning after pre-training on I2E-ImageNet.
  Transfer-II denotes fine-tuning after pre-training on I2E-CIFAR10.
  The transfer-II result of 92.5\% establishes a new state-of-the-art on real-world CIFAR10-DVS, demonstrating the effectiveness of the proposed sim-to-real training paradigm.}
    \label{table_cifar}
\end{table}

\subsection{Performance on I2E-Generated Datasets}
A key advantage of I2E's real-time nature is its compatibility with standard on-the-fly data augmentation pipelines, a technique precluded by the static nature of previous event datasets.
To quantify this benefit, two baseline conditions were evaluated.
In Baseline-I, only minimal augmentation, random horizontal flipping, was used.
In Baseline-II, a full suite of standard augmentations such as random cropping was applied to the source images before I2E conversion.

As shown in Table \ref{table_imagenet} and Table \ref{table_cifar}, models trained on I2E data achieve state-of-the-art results.
On I2E-ImageNet, MS-ResNet34 (Baseline-II) reaches 60.50\% accuracy, surpassing the best prior result on other event-based ImageNet datasets by over 8\%.
The dramatic performance increase from Baseline-I to Baseline-II across all datasets demonstrates that I2E is not only capable of generating high-quality event data but also uniquely enables the modern training strategies required to unlock the full potential of deep SNNs.

\subsection{Transfer Learning: A New Paradigm for SNNs}
The most significant contribution of this work is the establishment of a practical and effective pre-training paradigm for SNNs.
By providing a virtually unlimited source of low-cost, high-quality synthetic event data, I2E enables robust pre-training for subsequent fine-tuning on smaller, real-world event datasets, thereby addressing the critical data scarcity problem in neuromorphic engineering.

\paragraph{Transferability across I2E datasets}
First, the effectiveness of transfer learning within the I2E ecosystem was established.
As shown in Table~\ref{table_cifar}, a model pre-trained on the large-scale I2E-ImageNet dataset and then fine-tuned on I2E-CIFAR demonstrates significant performance gains.
Accuracy on I2E-CIFAR10 improves from 89.23\% to 90.86\%, while on the more challenging I2E-CIFAR100, accuracy sees a substantial boost from 60.68\% to 64.53\%.
This confirms that features learned on I2E-ImageNet are general and transferable to other I2E-generated tasks.

\begin{table}[t]
    \centering
    \begin{tabular}{p{4.3em} p{4.1em}<{\centering} p{3.8em}<{\centering} p{3.4em}<{\centering} p{2.3em}<{\centering}}
        \toprule[1.5pt]
        {Method} & \makecell{Dynamic \\ Threshold} & \makecell{Random \\ Selection}  & \makecell{Random \\ Crop} & \makecell{Acc. \\ \%}\\
        \midrule[1pt]
        {ablation-1} & $\times$ & $\times$ & $\times$ & 47.22 \\
        \midrule[1pt]
        {baseline-I} & $\checkmark$ & $\times$ & $\times$ & 48.30 \\
        \midrule[1pt]
        {ablation-2} & $\checkmark$ & $\checkmark$ & $\times$ & 49.01 \\
        \midrule[1pt]
        {baseline-II} & $\checkmark$ & $\checkmark$ & $\checkmark$ & 57.97 \\
        \bottomrule[1.5pt]
    \end{tabular}
    \caption{Ablation study of I2E components on I2E-ImageNet with ResNet18.
  Dynamic thresholding, random selection, and compatibility with standard augmentations are all essential for achieving optimal performance.}
    \label{table_ablation}
\end{table}
\begin{table}[t]
    \centering
    \begin{tabular}{p{3.7em} p{1.9em}<{\centering} p{1.9em}<{\centering} p{1.9em}<{\centering} p{1.9em}<{\centering} p{1.9em}<{\centering} p{1.9em}<{\centering}}
        \toprule[1.5pt]
        \multirow{2.2}{*}{Dataset} & \multicolumn{6}{c}{Acc.\%}\\
        \cline{2-7}
         & {$\alpha\beta\gamma$} & {$\alpha\gamma\beta$} & {$\beta\alpha\gamma$} & {$\beta\gamma\alpha$} & {$\gamma\alpha\beta$} & {$\gamma\beta\alpha$}\\
        \midrule[1pt]
        {CIFAR10} & {87.96} & {88.94} & {87.36} & {88.88} & \textbf{89.23} & {88.60}\\
        \midrule[1pt]
        {CIFAR100} & {56.10} & {59.43} & {55.11} & {59.25} & \textbf{60.68} & {60.12}\\
        \bottomrule[1.5pt]
    \end{tabular}
    \caption{Ablation on the timestep processing order.
  Presenting groups with higher event rates first (the $\gamma\alpha\beta$ sequence) consistently yields the best performance.}
    \label{table_order}
\end{table}

\paragraph{Bridging the sim-to-real gap}
The key experiment involves transferring knowledge from I2E-generated data to a real-world DVS sensor dataset.
A model was pre-trained on synthetic I2E-CIFAR10 data and then fine-tuned on the real-world CIFAR10-DVS dataset.
The result, as shown in Table \ref{table_cifar}, achieves a new state-of-the-art accuracy of 92.5\%, outperforming the previous best by a remarkable 7.7\%.
This successful sim-to-real transfer is a crucial finding, as it demonstrates that the event streams produced by I2E serve as a high-fidelity proxy for real sensor data.
It validates a powerful new workflow for the field: leverage vast static image libraries to pre-train robust SNNs via I2E, and then fine-tune them with a limited amount of costly, real-world DVS data to achieve state-of-the-art performance.
This paradigm mitigates the data acquisition bottleneck that has long hindered progress in neuromorphic computing.

\begin{table}[t]
    \centering
    \begin{tabular}{p{3.0em} p{3.0em}<{\centering} p{3.0em}<{\centering} p{2.5em}<{\centering}}
        \toprule[1.5pt]
        \multirow{1}{*}{Method} & {Channel} & Timestep & {Acc.\%}\\
        \midrule[1pt]
        {RGB} & $3$ & $4$ & 65.68 \\
        \midrule[1pt]
        {V} & $1$ & $4$ & 62.21 \\
        \midrule[1pt]
        {Event} & $2$ & $8$ & 59.28 \\
        \bottomrule[1.5pt]
    \end{tabular}
    \caption{Analysis of performance impact from RGB-to-Value conversion on ImageNet.}
    \label{table_loss}
\end{table}

\subsection{Ablation Studies and Analysis}
To validate the design choices of the I2E algorithm, a series of ablation studies and analyses were conducted.

\paragraph{Impact of algorithmic components}
The ablation study on ImageNet, as shown in Table \ref{table_ablation}, confirms the importance of I2E's core components by building the algorithm from the ground up.
Starting with a naive conversion (fixed threshold), the model achieves only 47.22\% accuracy.
Introducing the dynamic threshold stabilizes the event rate and improves performance to 48.30\%.
Adding the random selection provides essential data augmentation, further boosting accuracy to 49.01\%.
Finally, enabling standard augmentations such as random cropping, which is only possible due to I2E's real-time nature, provides the largest benefit, increasing accuracy to 57.97\%.
This highlights the synergy between the algorithm's design and modern training practices.

\paragraph{Analysis of timestep order}
The sequence in which event frames are processed affects performance.
The eight motion vectors were categorized into three groups ($\alpha, \beta, \gamma$) based on the magnitude of their modulus, as Figure \ref{fig_direction}, which correlates with the resulting event rate.
As shown in Table \ref{table_order}, ordering the groups to present frames with higher event rates first (the $\gamma\alpha\beta$ sequence) consistently yields the best performance on both CIFAR-10 and CIFAR-100.
Corresponding to the eight timesteps, the best order is: e, f, g, h, a, b, c, d.

\begin{figure}[t]
  \centering
  \includegraphics[width=0.75\linewidth]{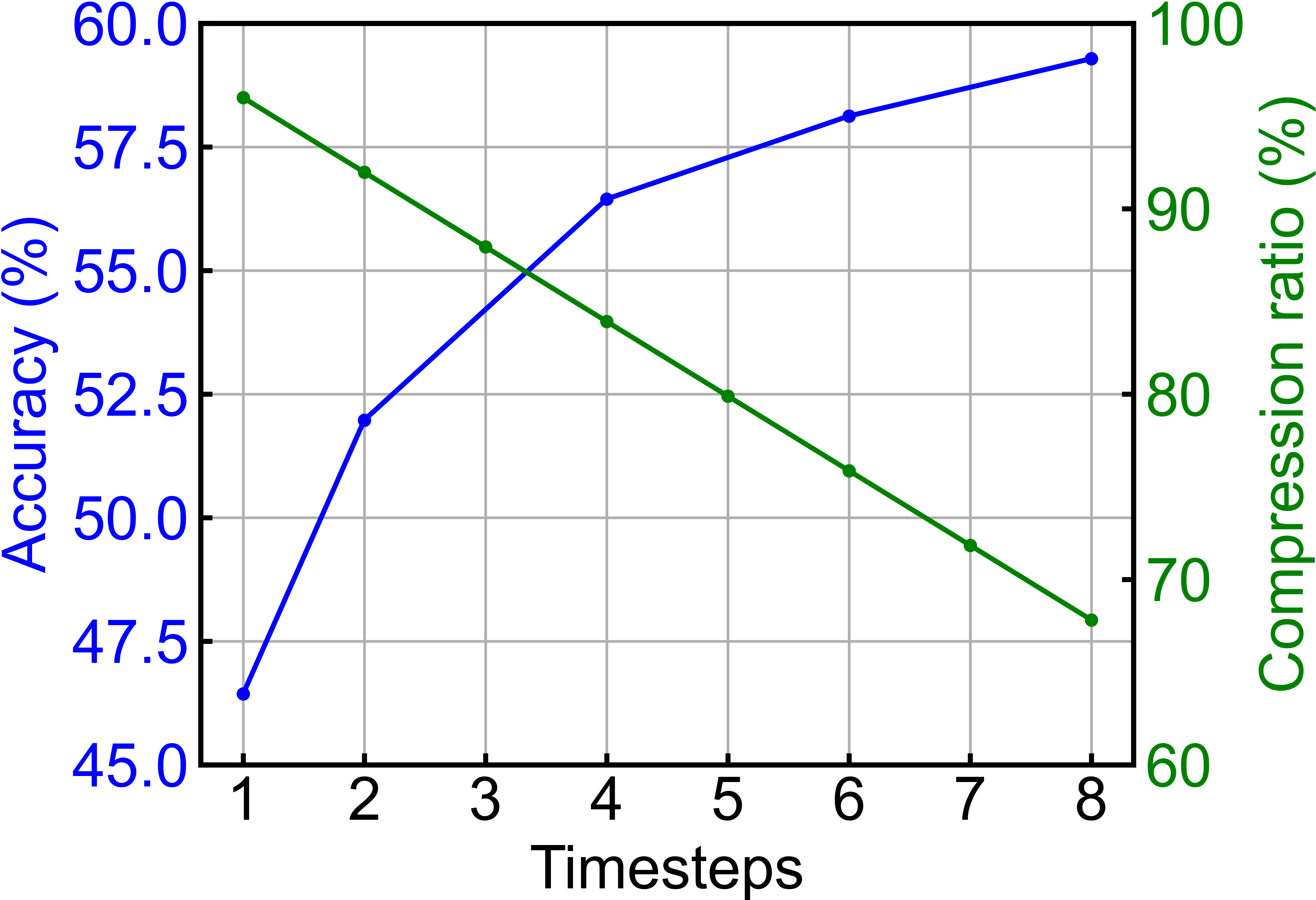}
  \caption{Trade-off between timesteps, accuracy, and data compression on ImageNet.
  Using more timesteps improves accuracy at the cost of a lower data compression ratio.}
  \label{fig_rate}
\end{figure}

\paragraph{Analysis of conversion loss and timesteps}
An analysis was conducted to quantify information loss during conversion.
As shown in Table~\ref{table_loss}, converting an RGB image to a single-channel Value map results in a performance drop from 65.68\% to 62.21\%, defining a practical upper bound for the event-based approach.
The final I2E-trained model achieves 59.28\%, indicating that while the conversion is highly effective, the inherent sparsity of events still introduces a minor performance trade-off.
Furthermore, the number of timesteps can be adjusted to balance accuracy and data compression as shown in Figure~\ref{fig_rate}.
Reducing the timesteps to just 2 still yields an accuracy of 51.97\% on ImageNet, which is competitive with prior work, while increasing the data compression ratio to 91.95\%.

%==================================================== Discussion ============================================================================
\section{Discussion and Limitations}
\label{sec:discussion}

The primary implication of the I2E is the establishment of a practical and highly effective pre-train, then fine-tune workflow for the neuromorphic domain.
This approach directly addresses the data scarcity and quality bottleneck that has long constrained SNN research.
The successful sim-to-real transfer experiment, in which a model pre-trained on synthetic I2E data achieved state-of-the-art performance on the real-world CIFAR10-DVS dataset, strongly validates this paradigm.
It demonstrates that I2E-generated data serves as a high-fidelity proxy for physical sensor data, effectively decoupling SNN model development from the slow and costly process of hardware-based data acquisition.

This development lowers the barrier to entry for SNN research and development.
By enabling ubiquitous, low-cost RGB cameras to function as effective event-based sensors through a software layer, I2E makes the design of energy-efficient SNNs more economically and logistically viable.
Furthermore, the algorithm's real-time nature unlocks new research avenues.
The systematic exploration of data augmentation strategies for event streams, a critical area for improving generalization, was previously inaccessible due to the static nature of existing event datasets.
While this study focused on validating the I2E paradigm for classification, extending this pre-training workflow to other complex tasks, such as detection and segmentation, or other event-driven tasks, remains a key direction for future work.

% \section{Discussion}
% \label{sec:discussion}

% The primary implication of the I2E framework is the establishment of a practical and highly effective pre-train, then fine-tune workflow for the neuromorphic domain.
% This approach directly addresses the data scarcity and quality bottleneck that has long constrained SNN research.
% The successful sim-to-real transfer experiment—in which a model pre-trained on synthetic I2E data achieved state-of-the-art performance on the real-world CIFAR10-DVS dataset—strongly validates this paradigm.
% It demonstrates that I2E-generated data serves as a high-fidelity proxy for physical sensor data, effectively decoupling SNN model development from the slow and costly process of hardware-based data acquisition.

% This development lowers the barrier to entry for SNN research and development.
% By enabling ubiquitous, low-cost RGB cameras to function as effective event-based sensors through a software layer, I2E makes the design of energy-efficient SNNs more economically and logistically viable.
% Furthermore, the algorithm's real-time nature unlocks new research avenues.
% The systematic exploration of data augmentation strategies for event streams—a critical area for improving generalization—was previously inaccessible due to the static nature of existing event datasets.
% The I2E algorithm, therefore, provides not just a tool but a foundational platform for future innovation in neuromorphic computing.

\section{Conclusion}
\label{sec:conclusion}

This work introduced I2E, an algorithmic framework that resolves a critical data bottleneck for SNNs by converting static images into high-fidelity event streams in real-time.
The method's efficiency, which is orders of magnitude faster than prior approaches, uniquely enables the use of modern on-the-fly data augmentation pipelines for SNN training.
The quality of the generated data was demonstrated by training a deep SNN on the new I2E-ImageNet dataset to a state-of-the-art accuracy of 60.50\%.
Critically, this work established a powerful sim-to-real paradigm by pre-training a model on synthetic I2E data and fine-tuning it on the real-world CIFAR10-DVS dataset, achieving an unprecedented accuracy of 92.5\%.
By open-sourcing the algorithm and datasets, this research provides the community with an essential toolkit to bridge the data gap, accelerating the development of high-performance, practical neuromorphic systems.
This work thus paves the way for deploying SNNs in complex, real-world applications where both high performance and extreme energy efficiency are required.

% =========================================================================

\section{Acknowledgments}
This work was supported in part by the Scientific and Technological Innovation (STI) 2030-Major Projects under Grant 2022ZD0209700, in part by Sichuan Science and Technology Program under Grant 2024ZDZX0001 and Grant 2024ZYD0253, and in part by Shenzhen Natural Science Foundation under JCYJ20250604180428038.

\bigskip
% \small
\bibliography{aaai2026}

% ======================================== Appendix ====================================
\newpage
\appendix
\section{Spiking Neural Networks}
The fundamental computational unit of a spiking neural network (SNN) is the spiking neuron.
The dynamics of a spiking neuron are governed by a discrete-time process that unfolds in three distinct stages: membrane potential integration, spike generation, and potential post-spike reset.
This process can be formulated as follows:
\begin{equation}
H[t]=f(V[t-1], X[t]),
\label{equ_dynamics}
\end{equation}
\begin{equation}
S[t]=\Theta(H[t]-V_{th}),
\label{equ_spike}
\end{equation}
\begin{equation}
V[t]=H[t](1-S[t])+V_{reset}S[t],
\label{equ:reset}
\end{equation}
Here, at each timestep $t$, $X[t]$ represents the input current to the neuron.
The function $f(\cdot)$ in Equation \ref{equ_dynamics} integrates this input with the previous state's membrane potential $V[t-1]$, to produce an intermediate potential $H[t]$.
The neuron generates a binary output spike, $S[t] \in \{0, 1\}$, if this potential $H[t]$ exceeds a predefined firing threshold $V_{th}$, which set to $1$.
This firing mechanism is modeled by the Heaviside step function $\Theta(\cdot)$, as shown in Equation \ref{equ_spike}, where $\Theta(x)=1$ for $x>0$ and is $0$ otherwise.
Following a spike ($S[t]=1$), the neuron's membrane potential $V[t]$ is reset to $V_{reset}=0$.
Otherwise, it retains the value of $H[t]$, as Equation \ref{equ:reset}.

The specific computational behavior of the neuron is determined by the choice of the integration function $f(\cdot)$.
The integration dynamics of the widely used Leaky Integrate-and-Fire (LIF) neuron are given by:
\begin{equation}
H[t]_{LIF} = V[t-1] - \frac{1}{\tau}(V[t-1] - V_{reset}) + X[t]
\label{no_decay_input}
\end{equation}
where $\tau$ is the membrane time constant, which controls the rate of potential decay.
We set $\tau=2$ in our experiments.
For constructing deep residual networks, we adopt the Membrane-Shortcut ResNet (MS-ResNet) architecture, where the residual connection is applied to the membrane potential $V[t]$ rather than the sparse spike signal $S[t]$.

\begin{figure}[t]
  \centering
  \includegraphics[width=0.8\linewidth]{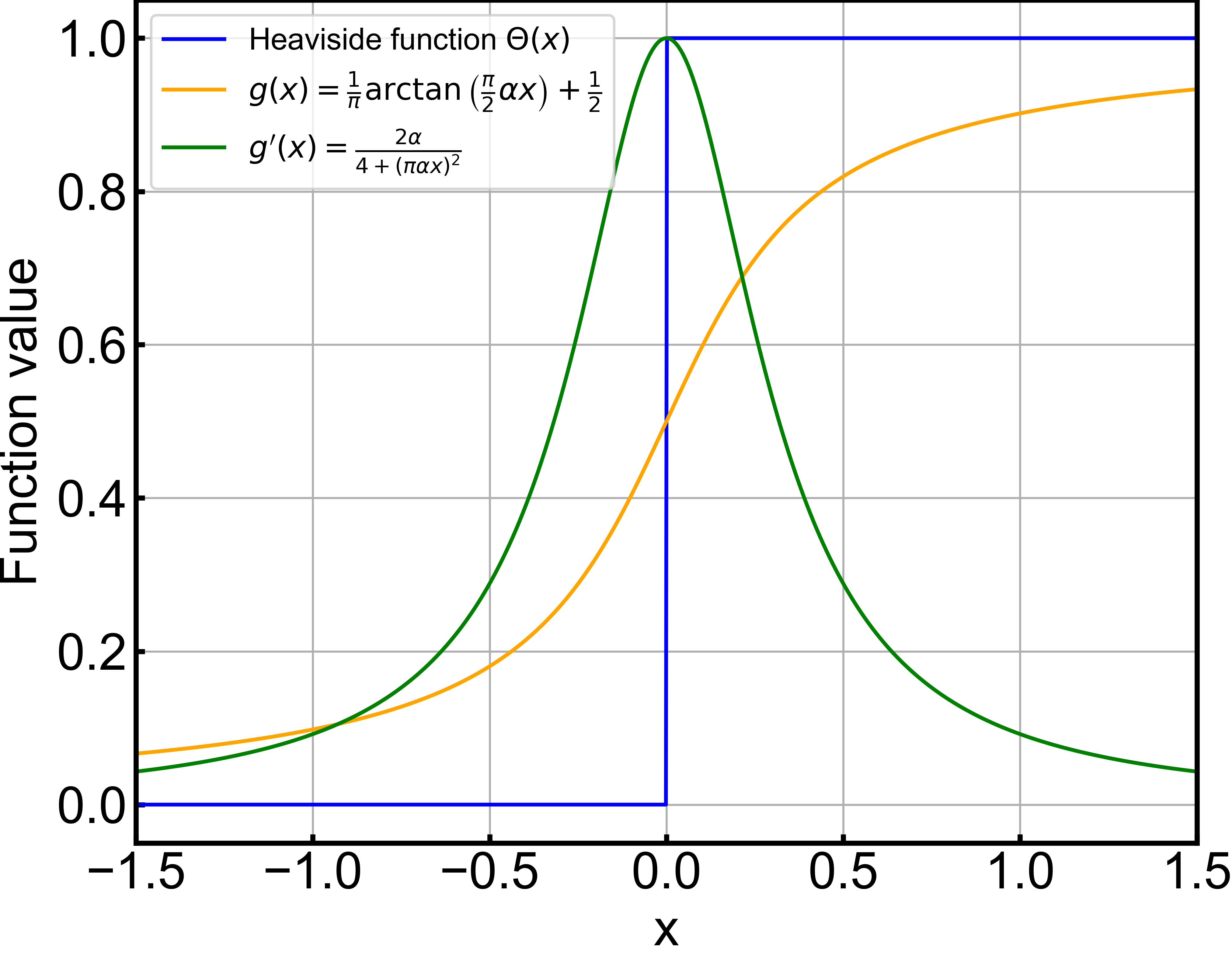}
  \caption{Visualization of the surrogate gradient method.
    The blue line represents the Heaviside step function used in the forward pass.
    The orange line shows the smooth arctan-based surrogate function $g(x)$.
    The green line illustrates the derivative of the surrogate function $g'(x)$, which is used to compute gradients during the backward pass.}
  \label{figs1}
\end{figure}

A significant challenge in training SNNs is the non-differentiable nature of the Heaviside function used for spike generation.
To enable gradient-based optimization, we employ the surrogate gradient method.
This approach replaces the true gradient of $\Theta(\cdot)$ during the backward pass with the gradient of a smooth, differentiable surrogate function.
Specifically, we use an arctan-based surrogate, defined as:
\begin{equation}
g(x) = \frac{1}{\pi} \arctan(\frac{\pi}{2}\alpha x) + \frac{1}{2}
\end{equation}
Its derivative, which is used in the backward pass, is:
\begin{equation}
g'(x) = \frac{2\alpha}{4+(\pi\alpha x)^2}
\end{equation}
The hyperparameter $\alpha$ controls the steepness of the surrogate function, effectively modulating the sharpness of the approximated gradient.
We set $\alpha=2$.
A visual comparison of the Heaviside step function, its surrogate, and the surrogate's gradient is provided in Figure \ref{figs1}.

\section{Detailed Energy Consumption Analysis}
This section provides a detailed breakdown of the theoretical energy consumption calculations referenced in the main text.
The fundamental energy-consuming operation in a conventional convolutional layer is the multiply-accumulate (MAC) operation, whereas in a spiking convolutional layer, it is the accumulate (AC) operation, triggered only upon receiving an input spike.
Following the prior work, we adopt energy cost values of $E_{\text{mac}} = 4.6 \, \text{pJ}$ for a 32-bit floating-point MAC operation and $E_{\text{ac}} = 0.9 \, \text{pJ}$ for an AC operation.

The energy consumption for an ANN layer ($E_{\text{ANN}}$) and an SNN layer ($E_{\text{SNN}}$) can thus be modeled as:
\begin{equation}
    E_{\text{ANN}} = N_{\text{ops}} \cdot E_{\text{mac}}
\end{equation}
\begin{equation}
    E_{\text{SNN}} = N_{\text{ops}} \cdot {fr} \cdot T \cdot E_{\text{ac}}
\end{equation}
where $N_{\text{ops}}$ is the total number of synaptic operations (MAC or AC) in the layer, $fr$ is the average firing rate of the neurons, and $T$ is the number of simulation timesteps.
The theoretical number of operations for a standard convolutional layer is calculated as:
\begin{equation}
    N_{\text{ops}} = K^2 \cdot C_{\text{in}} \cdot C_{\text{out}} \cdot H_{\text{out}} \cdot W_{\text{out}}
    \label{eq:ops_calc}
\end{equation}
Here, $K$ is the kernel size, $C_{\text{in}}$ and $C_{\text{out}}$ are the input and output channel counts, and $H_{\text{out}}$ and $W_{\text{out}}$ are the spatial dimensions of the output feature map.

To illustrate the practical implications, we analyze the first layer of a ResNet-style architecture processing inputs from the ImageNet dataset ($224 \times 224 \times 3$).
This layer has a kernel size $K=7$, stride 2, $C_{\text{in}}=3$, $C_{\text{out}}=64$, and produces an output of size $112 \times 112$.

For the conventional ANN, the energy cost is:
\begin{equation}
    E_{\text{ANN}} = (7^2 \cdot 3 \cdot 64 \cdot 112^2) \cdot 4.6 \, \text{pJ} \approx 543 \, \mu\text{J}
\end{equation}

For our proposed I2E-SNN, the first layer processes the output of the I2E module, which has $C_{\text{in}}=2$.
Assuming a typical firing rate ${fr}=5\%$ and $T=8$, the energy cost is:
\begin{equation}
    E_{\text{SNN}} = (7^2 \cdot 2 \cdot 64 \cdot 112^2) \cdot 0.05 \cdot 8 \cdot 0.9 \, \text{pJ} \approx 28 \, \mu\text{J}
\end{equation}

The I2E encoding itself functions as a lightweight convolutional layer with $C_{\text{in}}=1$, $C_{\text{out}}=8$ (for $T=8$), operating on the full input resolution.
Its energy cost is minimal:
\begin{equation}
    E_{\text{I2E}} = (1^2 \cdot 1 \cdot 8 \cdot 224^2) \cdot 0.9 \, \text{pJ} \approx 0.36 \, \mu\text{J}
\end{equation}

The total energy for the first effective layer of our model is $E_{\text{SNN}} + E_{\text{I2E}} \approx 28.68 \, \mu\text{J}$.
Compared to the standard ANN's first layer, this represents a $\mathbf{18.9\times}$ reduction in energy consumption.
This energy efficiency can be further improved, for instance, reducing the timestep to $T=2$ lowers the first layer's energy consumption to approximately $7.17 \, \mu\text{J}$, achieving a $\mathbf{75.7\times}$ reduction relative to the ANN baseline.

\section{Data Representation and Compression}

To facilitate diverse experimental setups and hardware targets, we converted the static ImageNet dataset into two distinct event-based formats.
This section details the structure of these formats and analyzes the resulting data compression.
The original ImageNet dataset, comprising static JPEG images, occupies 146 GB of storage.

\subsection{Dense Tensor Representation}
The first format is a dense tensor representation, designed for seamless integration with deep learning frameworks that expect structured, multi-dimensional arrays.
Each image is converted into a Boolean tensor with shape $[T, C, H, W]$, where:
\begin{itemize}
    \item $T$ is the number of simulation timesteps.
    \item $C$ represents the polarity channels (e.g., ON/OFF events), resulting in $C=2$.
    \item $H$ and $W$ are the spatial dimensions of the image.
\end{itemize}
For an ImageNet image processed with $T=8$ timesteps, this results in a tensor of shape $[8, 2, 224, 224]$.
Each tensor is stored as a separate \textit{.npz} file.
The completely converted dataset in this format, including both training and validation sets, occupies \textbf{47 GB}.

\subsection{Sparse Event-Stream Representation}
The second format mimics the native output of DVS.
It is a sparse, coordinate-list representation where only pixel changes are recorded.
Each image is converted into a list of events, where each event is a tuple $(t, p, x, y)$ representing:
\begin{itemize}
    \item $t$: The discrete timestep at which the event occurred.
    \item $p$: The polarity of the event (e.g., 0 for OFF, 1 for ON).
    \item $x, y$: The spatial coordinates of the pixel that fired.
\end{itemize}
This list is stored as a multi-column matrix of type \textit{uint8} in an \textit{.npz} file.
This event-based format is inherently more efficient for sparse data.
The total size of the ImageNet dataset converted to this format is \textbf{44 GB}.

\subsection{Compression Analysis}
Both event-based representations offer substantial data compression compared to the original 146 GB dataset.
The compression ratio is calculated as $1 - \frac{\text{Encoded Size}}{\text{Original Size}}$.

\begin{itemize}
    \item For the dense tensor format ($T=8$):
    \begin{equation}
        \text{Compression Ratio} = 1 - \frac{47 \, \text{GB}}{146 \, \text{GB}} \approx 67.81\%
    \end{equation}
    \item For the sparse event-stream format:
    \begin{equation}
        \text{Compression Ratio} = 1 - \frac{44 \, \text{GB}}{146 \, \text{GB}} \approx 69.86\%
    \end{equation}
\end{itemize}

Furthermore, the size of both formats scales linearly with the number of timesteps.
This allows for even greater compression when fewer timesteps are required.
For instance, in our ablation study using $T=2$, the storage requirement for the dense format would be reduced by a factor of four.
This yields a significantly higher compression ratio:
\begin{equation}
    \text{Compression Ratio} \approx 1 - \frac{47/4 \, \text{GB}}{146 \, \text{GB}} \approx 91.95\%
\end{equation}
This highlights the efficiency of event-based encoding, particularly for applications where a short temporal window is sufficient.

\begin{figure*}[!t]
  \centering
  \begin{subfigure}{0.2623\linewidth}
    \centering
    \includegraphics[width=\linewidth]{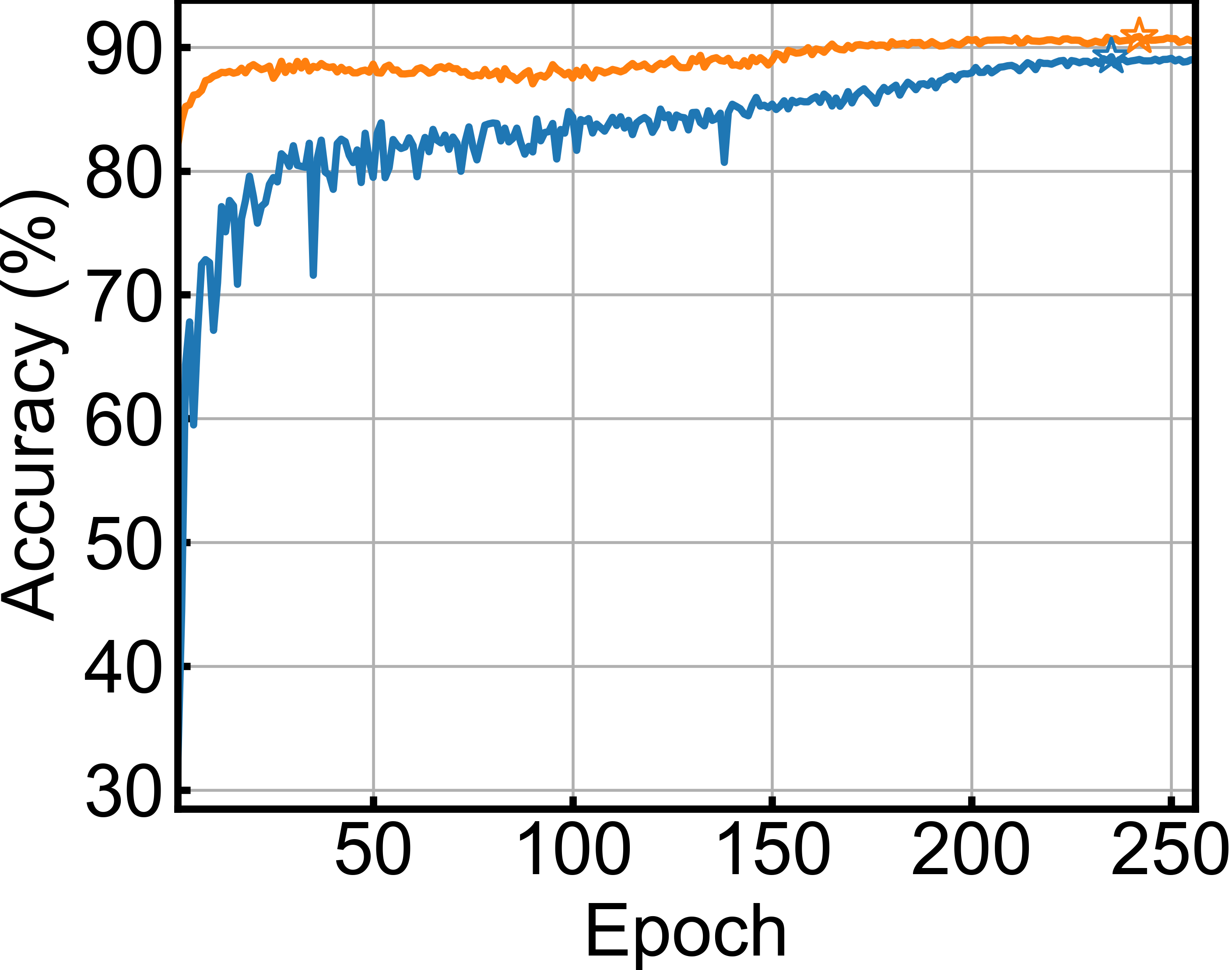}
    \caption{I2E-CIFAR10.}
    \label{figs1a}
  \end{subfigure}
  \begin{subfigure}{0.2444\linewidth}
    \centering
    \includegraphics[width=\linewidth]{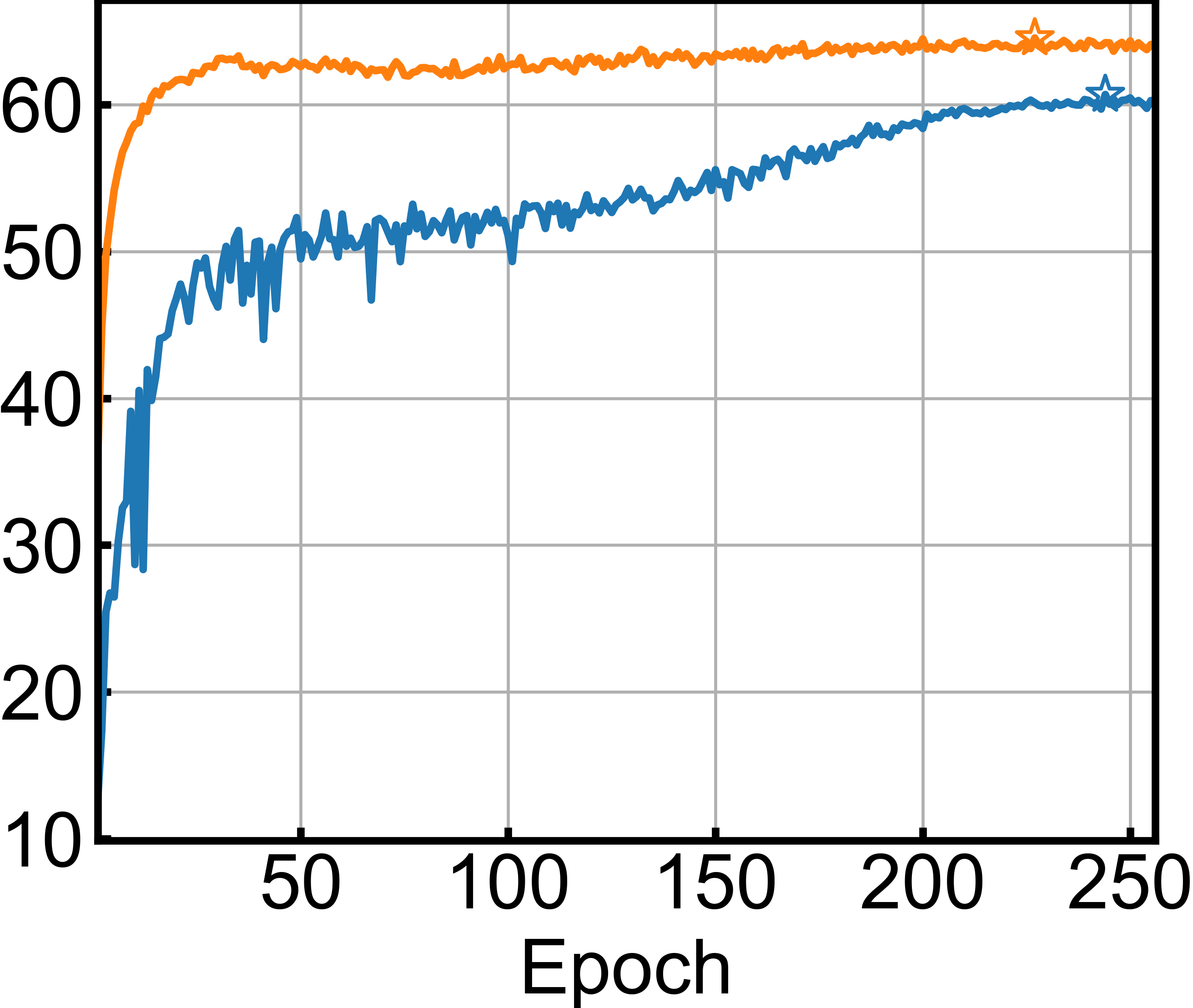}
    \caption{I2E-CIFAR100.}
    \label{figs1b}
  \end{subfigure}
  \begin{subfigure}{0.2366\linewidth}
    \centering
    \includegraphics[width=\linewidth]{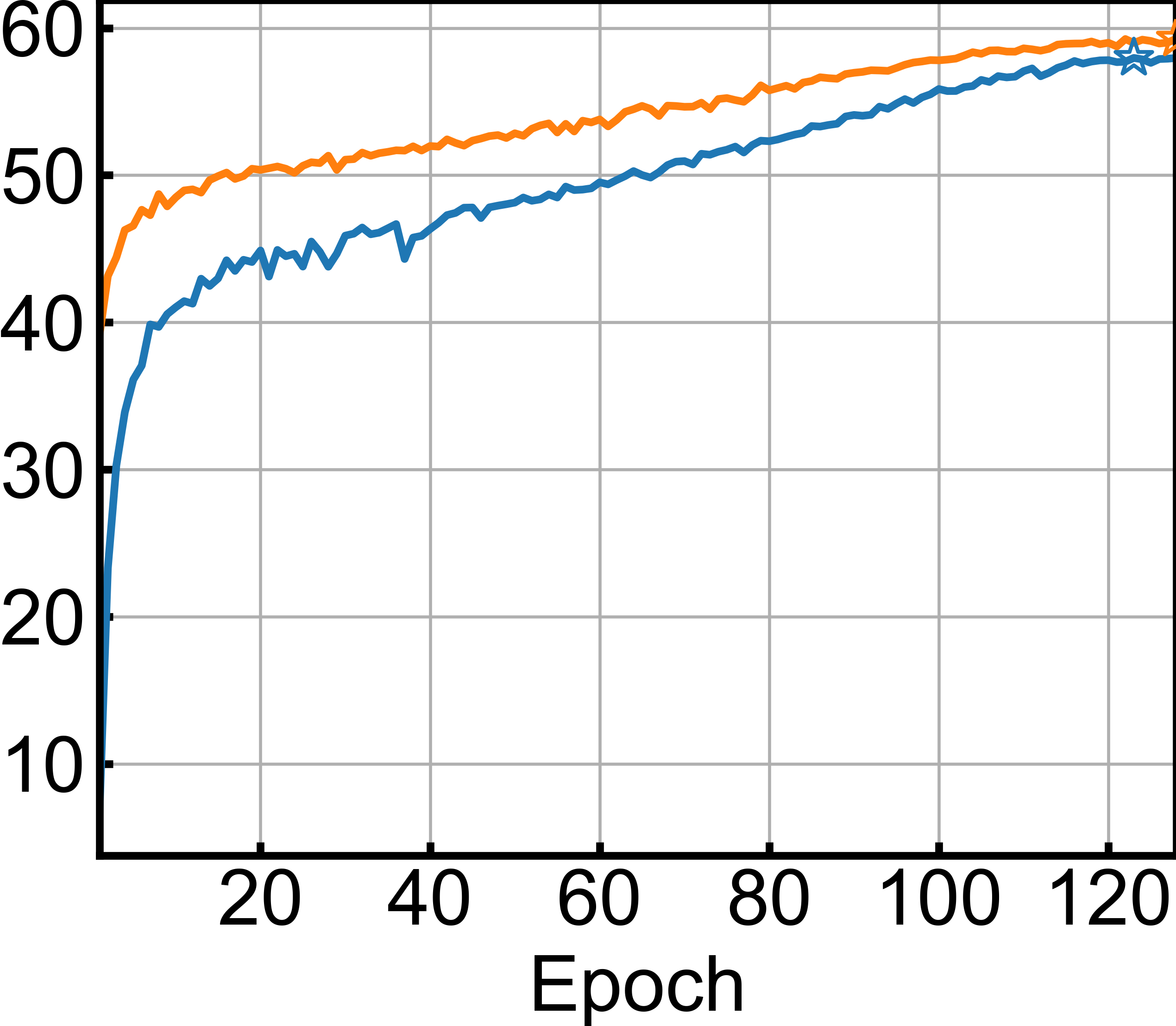}
    \caption{I2E-ImageNet.}
    \label{figs1c}
  \end{subfigure}
  \begin{subfigure}{0.2366\linewidth}
    \centering
    \includegraphics[width=\linewidth]{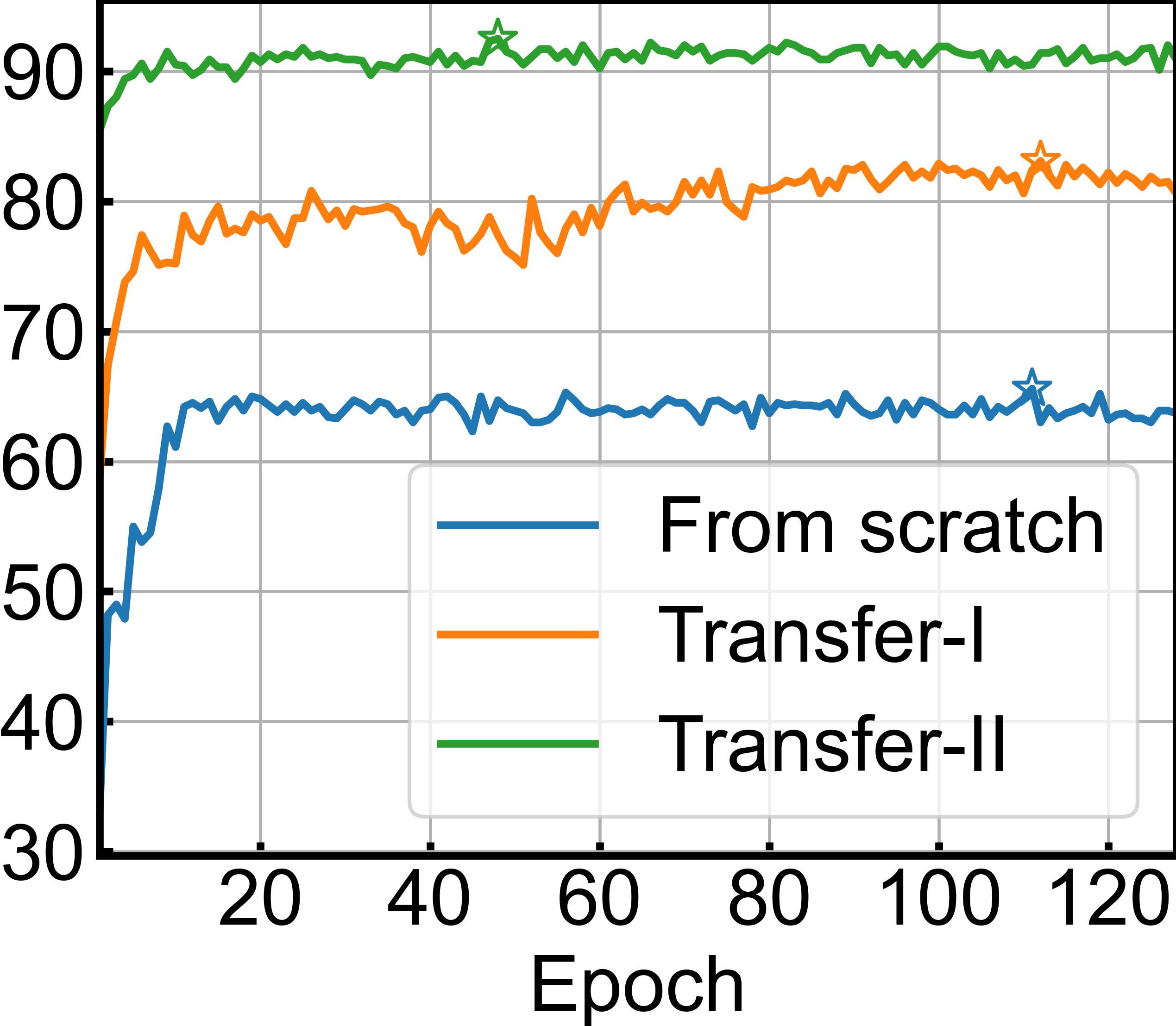}
    \caption{CIFAR10-DVS.}
    \label{figs1d}
  \end{subfigure}
  \caption{Validation accuracy curves on four datasets.
    Each plot compares the performance of models trained from scratch against those fine-tuned using transfer learning strategies.}
  \label{figs2}
\end{figure*}

\section{Experimental Details}
\subsection{Experimental Setup and Training Configuration}
All experiments were conducted on an Ubuntu 22.04 server equipped with two NVIDIA RTX 4090 GPUs.
To ensure reproducibility, all random seeds for data shuffling, initialization, and other stochastic processes were fixed to 2024.

We employed the SGD optimizer with a momentum of 0.9 for all training procedures.
The learning rate was decayed using a cosine annealing schedule.
Following common practice in SNN training to enhance stability and performance, the gradient flow through the reset mechanism (Equation \ref{equ:reset}) was detached during the backward pass.
The specific hyperparameters for each dataset and method are detailed in Table \ref{table_hyper1} (for training from scratch) and Table \ref{table_hyper2} (for fine-tuning).
\begin{table}[t]
    \centering
    \begin{tabular}{p{5.5em} p{3.7em}<{\centering} p{2.9em}<{\centering} p{2.4em}<{\centering} p{2.5em}<{\centering}}
        \toprule[1.5pt]
        {Dataset} & \makecell{Learning \\ Rate} & \makecell{Weight \\ Decay} & \makecell{Batch \\ Size} & {Epoch}\\
        \midrule[1pt]
        \multirow{1}{*}{I2E-CIFAR} & 0.1 & 2e-4 & 128 & 256\\
        % \cline{1-4}
        \multirow{1}{*}{ImageNet} & 0.1 & 2e-5 & 128 & 128\\
        \multirow{1}{*}{I2E-ImageNet} & 0.1 & 1e-5 & 128 & 128\\
        % \cline{1-4}
        \multirow{1}{*}{CIFAR10-DVS} & 0.1 & 1e-5 & 32 & 128\\
        \bottomrule[1.5pt]
    \end{tabular}
    \caption{Hyperparameters for training models from scratch.}
    \label{table_hyper1}
\end{table}
\begin{table}[t]
    \centering
    \begin{tabular}{p{5.5em} p{4.5em}<{\centering} p{3.7em}<{\centering} p{2.9em}<{\centering}}
        \toprule[1.5pt]
        {Dataset} & Method & \makecell{Learning \\ Rate} & \makecell{Weight \\ Decay}\\
        \midrule[1pt]
        \multirow{1}{*}{I2E-CIFAR} & transfer-I & 0.01 & 2e-4 \\
        \cline{2-4}
        \multirow{1}{*}{I2E-ImageNet} & transfer-I & 0.05 & 1e-5 \\
        \cline{2-4}
        \multirow{2}{*}{CIFAR10-DVS} & transfer-I & 0.01 & 1e-3 \\
         & transfer-II & 0.001 & 0 \\
        \bottomrule[1.5pt]
    \end{tabular}
    \caption{Hyperparameters for fine-tuning models.}
    \label{table_hyper2}
\end{table}

\subsection{Dataset-Specific Protocols}
\paragraph{CIFAR10-DVS}
This native event-based dataset contains 10,000 samples.
We designated 90\% of the data for the training set and the remaining 10\% for the validation set.
The raw event streams were integrated into dense frames over $T=8$ timesteps.
No data augmentation techniques were applied during the training process.

\paragraph{I2E-converted datasets}
For I2E-CIFAR and I2E-ImageNet, the static images were converted into event-based formats as described in the main text.
The training and fine-tuning followed the hyperparameter settings specified in Tables \ref{table_hyper1} and \ref{table_hyper2}.

\subsection{Transfer Learning and Fine-Tuning Strategies}
We investigated several transfer learning scenarios to evaluate the efficacy of pre-training.
The specific fine-tuning hyperparameters are listed in Table \ref{table_hyper2}.

\begin{itemize}
    \item \textbf{I2E-ImageNet:} The model was first pre-trained on the standard static ImageNet dataset.
    The first convolutional layer's weights were adapted to the dual-channel event input by removing the parameters corresponding to the third input channel of the original RGB data.
    The model was then fine-tuned on the I2E-ImageNet dataset.
    \item \textbf{I2E-CIFAR:} The models were pre-trained on the large-scale I2E-ImageNet dataset and subsequently fine-tuned on the I2E-CIFAR datasets.
    \item \textbf{CIFAR10-DVS:} We explored two pre-training strategies:
    \begin{itemize}
        \item \textbf{Transfer-I:} Pre-trained on I2E-ImageNet.
        \item \textbf{Transfer-II:} Pre-trained on I2E-CIFAR10.
    \end{itemize}
\end{itemize}
As illustrated by the training accuracy curves in Figure \ref{figs2}, pre-training consistently provided a significant performance improvement across all evaluated datasets.

\begin{table}[t]
    \centering
    \begin{tabular}{p{3em} p{4.5em}<{\centering} | p{3em}<{\centering} p{4.5em}<{\centering}}
        \toprule[1.5pt]
        {Group} & Time & Group & Time\\
        \midrule[1pt]
        {$\alpha\beta\gamma$} & abcdefgh & {$\alpha\gamma\beta$} & abefghcd \\
        {$\beta\alpha\gamma$} & cdabefgh & {$\beta\gamma\alpha$} & cdefghab \\
        {$\gamma\alpha\beta$} & efghabcd& {$\gamma\beta\alpha$} & efghcdab \\
        \bottomrule[1.5pt]
    \end{tabular}
    \caption{Mapping of group orders to temporal sequence.}
    \label{table_s3}
\end{table}

\subsection{Ablation Study Details}
For our ablation studies, the default simulation timestep was set to $T=4$ unless specified otherwise.

\paragraph{Timestep reduction}
In the experiment analyzing the impact of fewer timesteps, we constructed datasets with $T<8$ by systematically selecting the first $n$ timesteps from the fully generated 8-timestep data.

\paragraph{Temporal order}
In the experiments studying the influence of temporal sequence, the 8 timesteps were divided into three groups ($\alpha, \beta, \gamma$).
Table \ref{table_s3} details the specific mapping between group permutations and the resulting temporal order of the timesteps.

\subsection{Visualization of the I2E Conversion Process}
\label{sec:vis_example}

\begin{figure*}[!t]
    \centering
    \begin{subfigure}{0.4\linewidth}
        \centering
        \includegraphics[width=0.5\textwidth]{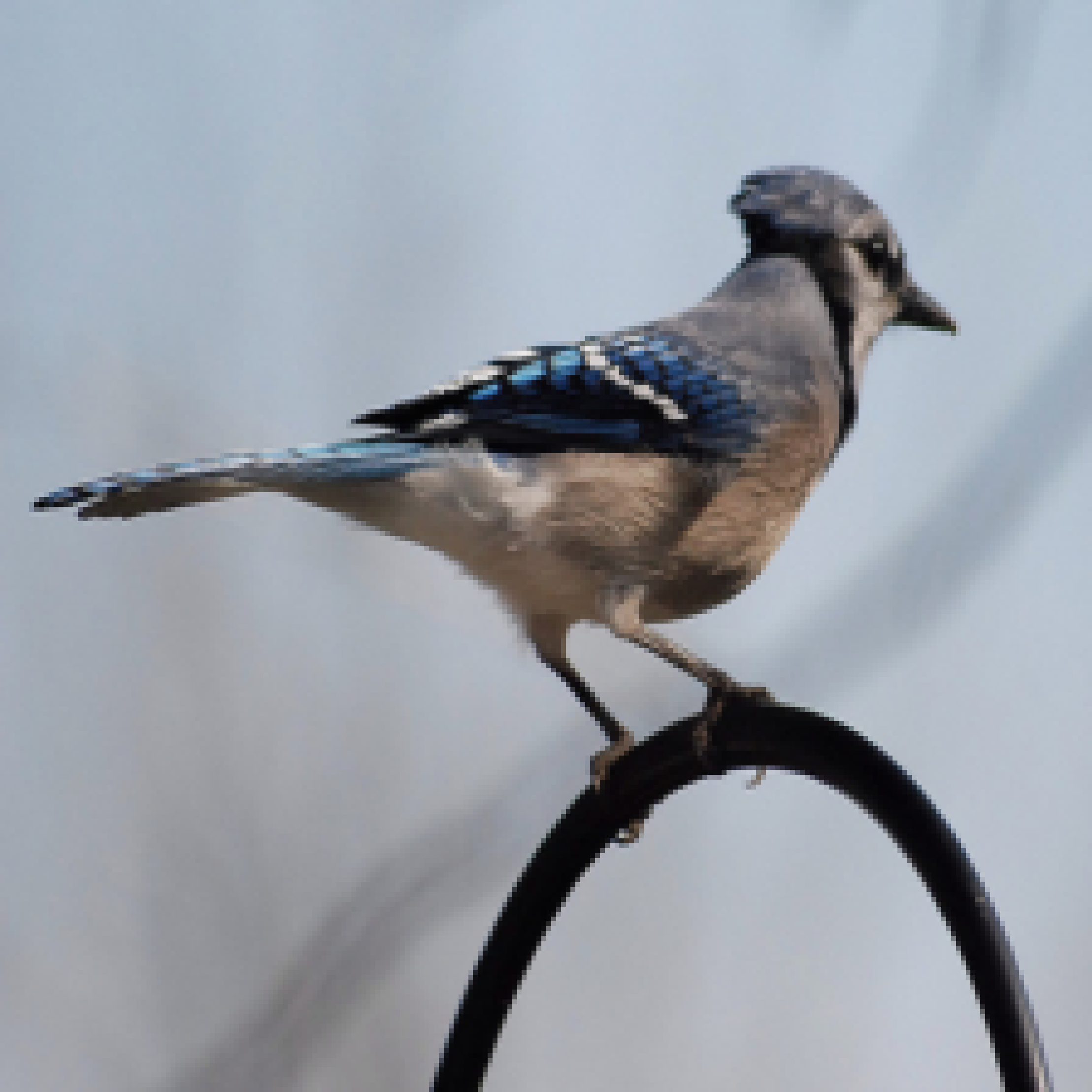}
        \caption{Original RGB image.}
        \label{fig:rgb}
    \end{subfigure}
    \centering
    \begin{subfigure}{0.4\linewidth}
        \centering
        \includegraphics[width=0.5\textwidth]{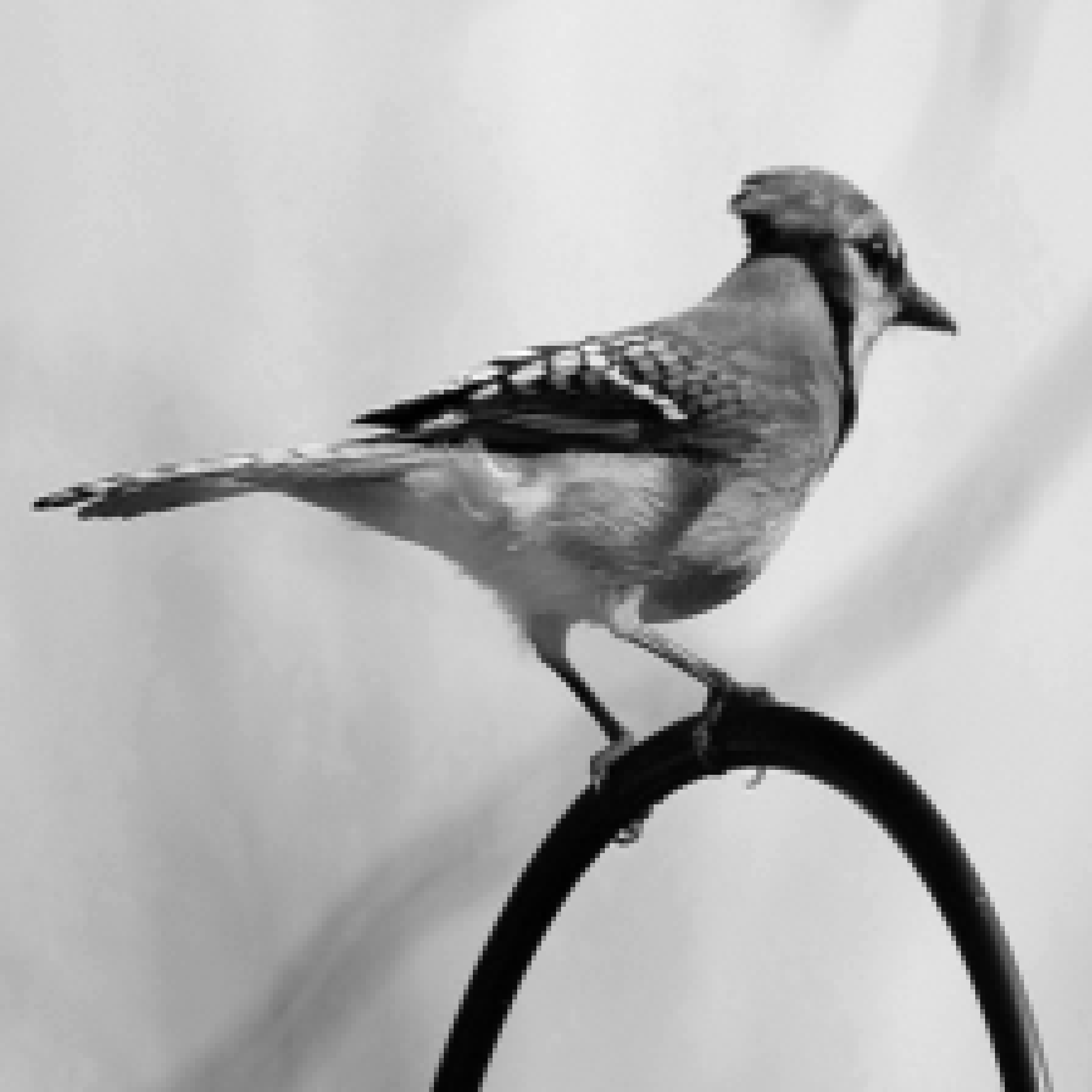}
        \caption{Intensity V-map.}
        \label{fig:vmap}
    \end{subfigure}

    \centering
    \begin{subfigure}{\linewidth}
        \centering
        \includegraphics[width=\textwidth]{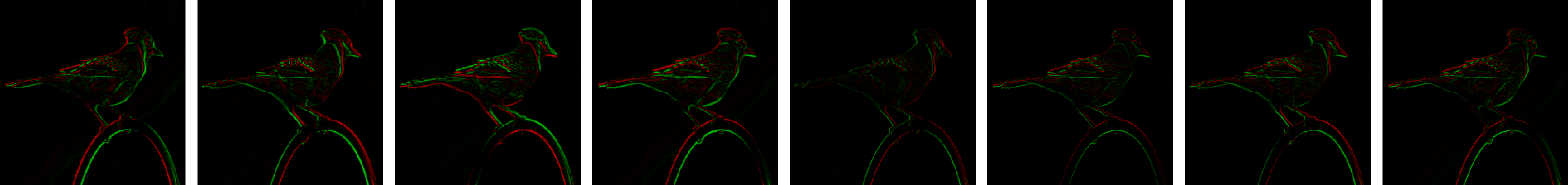}
        \caption{Stage 2: Float-value maps after convolution (T=8, C=2).}
        \label{fig:float_events}
    \end{subfigure}

    \centering
    \begin{subfigure}{\linewidth}
        \centering
        \includegraphics[width=\textwidth]{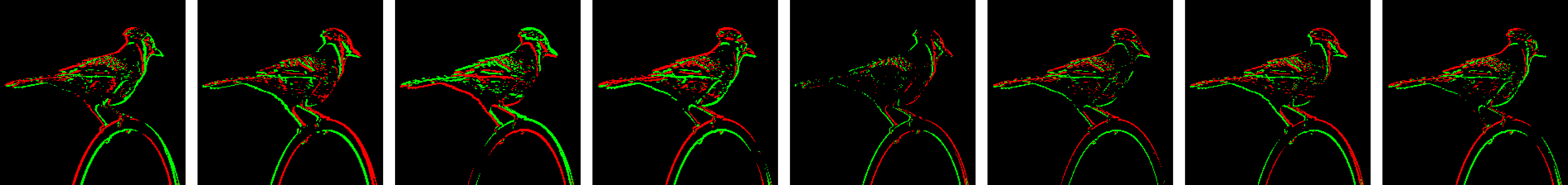}
        \caption{Stage 3: Final binary spikes after thresholding (T=8, C=2).}
        \label{fig:binary_spikes}
    \end{subfigure}

    \caption{A complete visualization of the I2E conversion process.
    The figure illustrates the complete data flow for a single sample, progressing from (a) original RGB image to (b) its intensity V-map, then to (c) the intermediate float-value maps, and finally (d) the binary event stream.}
    \label{fig:i2e_process}
\end{figure*}

\begin{figure*}[!t]
    \centering
    \begin{subfigure}{0.4\linewidth}
        \centering
        \includegraphics[width=0.5\textwidth]{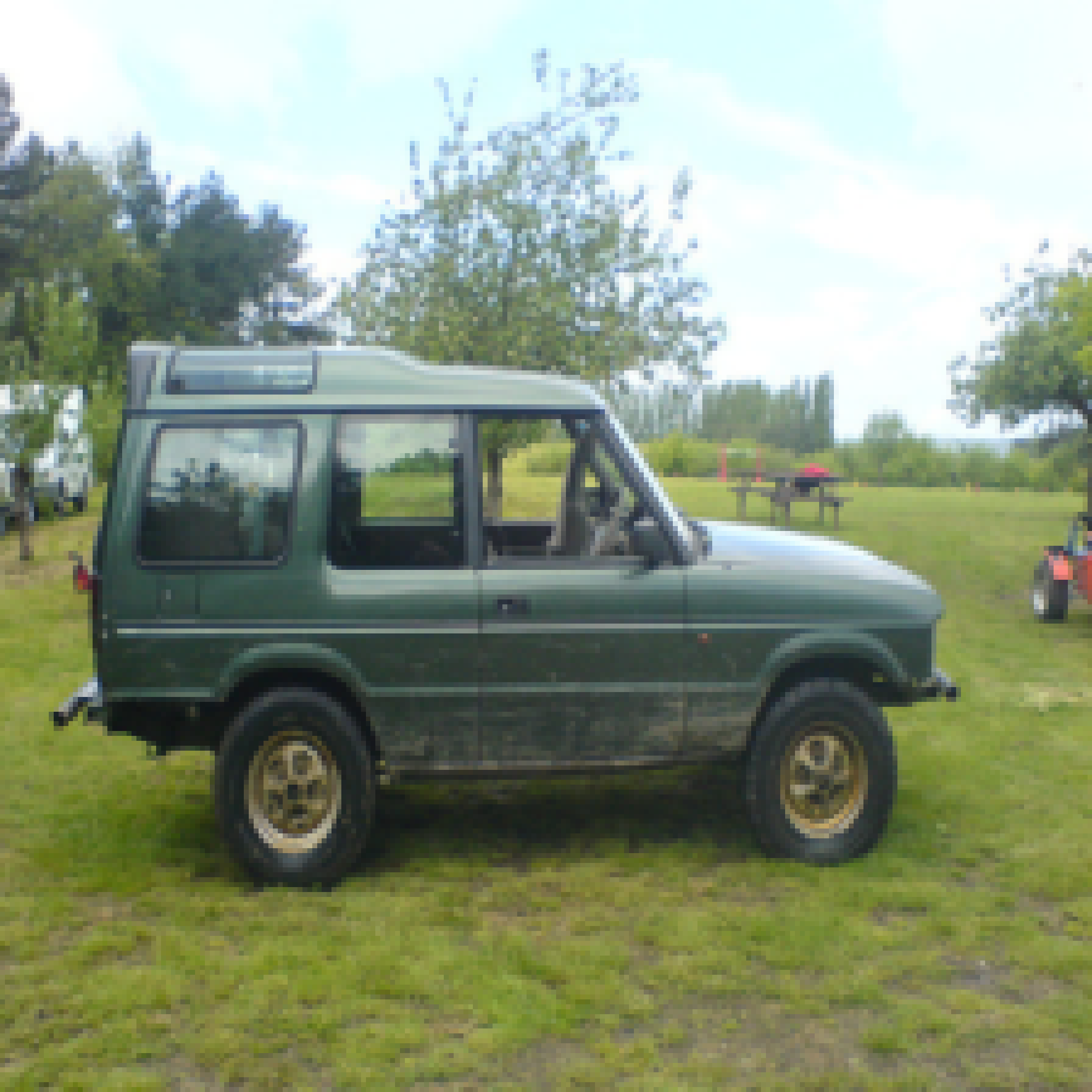}
        \caption{Original RGB image.}
        % \label{fig:rgb}
    \end{subfigure}
    \centering
    \begin{subfigure}{0.4\linewidth}
        \centering
        \includegraphics[width=0.5\textwidth]{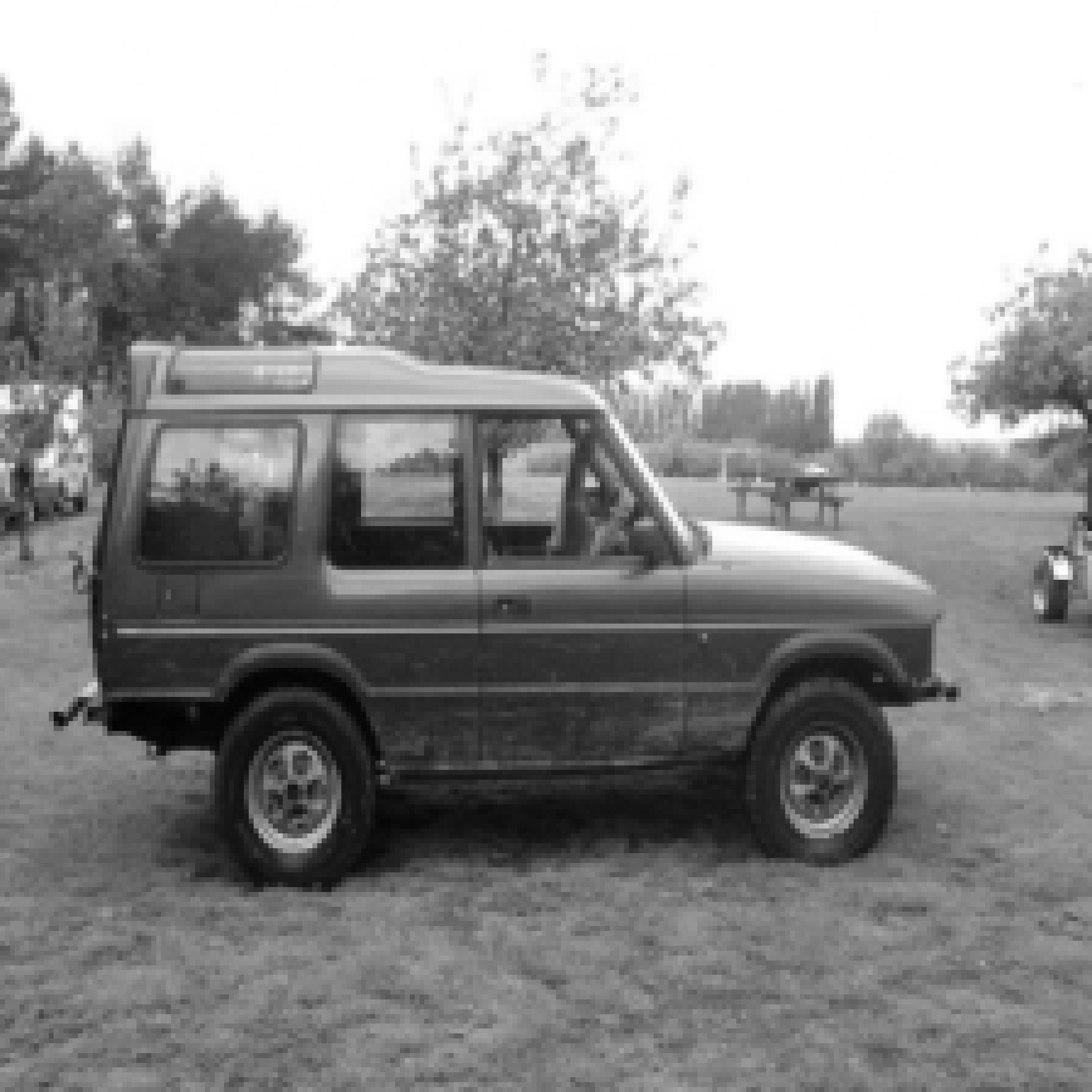}
        \caption{Intensity V-map.}
        % \label{fig:vmap}
    \end{subfigure}

    \centering
    \begin{subfigure}{\linewidth}
        \centering
        \includegraphics[width=\textwidth]{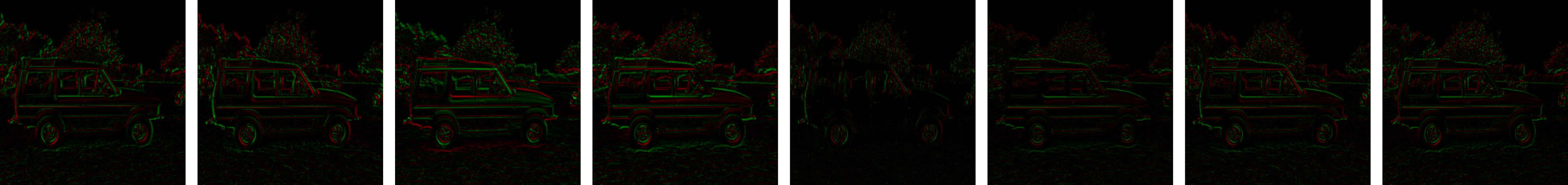}
        \caption{Stage 2: Float-value maps after convolution (T=8, C=2).}
        % \label{fig:float_events}
    \end{subfigure}

    \centering
    \begin{subfigure}{\linewidth}
        \centering
        \includegraphics[width=\textwidth]{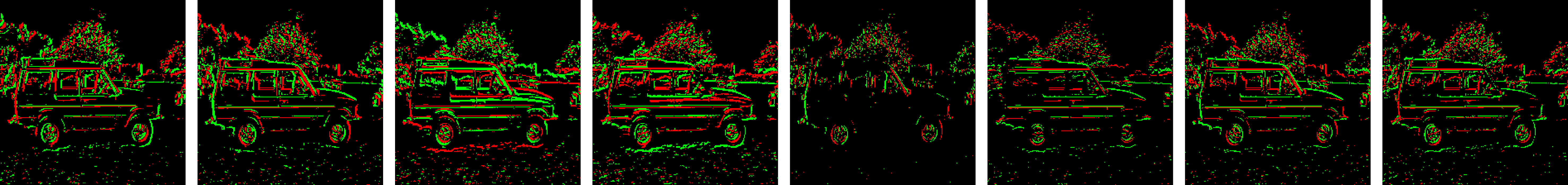}
        \caption{Stage 3: Final binary spikes after thresholding (T=8, C=2).}
        % \label{fig:binary_spikes}
    \end{subfigure}

    \caption{Another example.}
    \label{fig:i2e_process_}
\end{figure*}

To illustrate the I2E algorithm's pipeline, we provide a full conversion example for a sample image in Figure~\ref{fig:i2e_process} and \ref{fig:i2e_process_}.
This process clearly demonstrates the three main stages of converting a static image into a sparse event stream:

\begin{itemize}
    \item \textbf{Stage 1 (Row 1):} The original RGB image (a) is converted into a single-channel intensity V-map (b), which serves as the basis for subsequent calculations.
    \item \textbf{Stage 2 (Row 2):} The V-map is processed with 8 directional convolutional kernels, generating 8 timesteps of intensity change maps (c). The figure shows the float-point values for the two channels (ON/OFF), representing the magnitude and direction of brightness changes.
    \item \textbf{Stage 3 (Row 3):} The float-point values are compared against the adaptive threshold to generate the final binary event stream (d). This is the sparse data fed into the SNN.
\end{itemize}

\end{document}